\documentclass{article}
\usepackage{jfrExamplee}
\usepackage{graphicx}
\usepackage{setspace}
\usepackage{subfigure}
\usepackage{enumitem}
\usepackage{svg}
\usepackage{wasysym}
\usepackage{hyperref}
\usepackage{gensymb}
\usepackage{multirow}
\usepackage{booktabs}
\usepackage{notoccite}

\title{Autonomous Aerial Robot for High-Speed Search and Intercept Applications}

\author{Alejandro Rodriguez-Ramos,
        Adrian Alvarez-Fernandez\\
        \textbf{Hriday Bavle,
        Javier Rodriguez-Vazquez,
        Liang Lu}\\
        \textbf{Miguel Fernandez-Cortizas,
        Ramon A. Suarez Fernandez,
        Alberto Rodelgo}\\
        \textbf{Carlos Santos,
        Martin Molina,
        Luis Merino,
        Fernando Caballero and
        Pascual Campoy}
}

\begin{document}

\maketitle

\begin{abstract}
In recent years, high-speed navigation and environment interaction in the context of aerial robotics has become a field of interest for several academic and industrial research studies. In particular, Search and Intercept (SaI) applications for aerial robots pose a compelling research area due to their potential usability in several environments. Nevertheless, SaI tasks involve a challenging development regarding sensory weight, on-board computation resources, actuation design and algorithms for perception and control, among others. In this work, a fully-autonomous aerial robot for high-speed object grasping has been proposed. As an additional sub-task, our system is able to autonomously pierce balloons located in poles close to the surface. Our first contribution is the design of the aerial robot at an actuation and sensory level consisting of a novel gripper design with additional sensors enabling the robot to grasp objects at high speeds. The second contribution is a complete software framework consisting of perception, state estimation, motion planning, motion control and mission control in order to rapid- and robustly perform the autonomous grasping mission. Our approach has been validated in a challenging international competition and has shown outstanding results, being able to autonomously search, follow and grasp a moving object at 6 m/s in an outdoor environment.

\end{abstract}

\section{Introduction}
Research contributions in Unmanned Aerial Vehicles (UAVs) have considerably increased over the last decade, both in academic and industrial contexts \cite{gioioso2014flying,klingbeil2014towards,loianno2016estimation,ackerman2018skydio}. Indeed, these types of robotic systems, such as multirotor systems, provide a high maneuverability and an elevated potential level of autonomy in the field of complex missions. Furthermore,  the 6-Degrees Of Freedom (DOF) design of a multirotor allows for both precise and slow maneuvers, as well as fast and aggressive navigation or environment interaction, which can be considerably useful in the context of field robotics.

A notable amount of research has been carried out in the area of autonomous navigation or environment interaction (\textit{e.g.} object grasping). Stated autonomous tasks involve the development and coordination of several complex software components, as well as the integration of miniaturized sensors which provide usable rich information in order to successfully fulfill the endeavor. In particular, sensors such as RGB cameras or LIDARs are able to provide high-dimensional information of the environment which has to be efficiently processed on-board in order to perform the mission. In addition to the aforementioned challenges, high-speed environment navigation and/or interaction in the context of Search and Intercept (SaI) applications pose an enormous challenge due to the available on-board processing resources, sensor weight and algorithms to compute the high-dimensional information, efficiently and in a real time trend.

In this respect, regarding high-speed SaI applications, the complexity of the task increases with the required speed of the mission, since the algorithms have to be processed at a higher rate, and the dynamics of the system (\textit{e.g.} moments of inertia) show up accordingly. Normally, the robotic system is modeled for control, incurring several assumptions which simplify the design. These assumptions may not be valid as speed increases, since the non-linearities of the system are more notable, among others. Furthermore, the application of object interception with UAVs constitutes an increased challenge, due to constraints on actuator design and its effect on the dynamics of the system, as well as the required perception and control algorithms which enable the interaction maneuvers.

On this subject, a fully-autonomous UAV solution for SaI applications is proposed in this work. Although Global Positioning System (GPS) and Real Time Kinematic (RTK) systems were used in the outdoor scenario, the complete mission has been approached based on the RGB camera sensor as the main information source, which poses an additional challenge due to the environmental area size, conditions and amount information to be processed at high-speeds. In this study, the complete UAV system is described, as well as the actuator design for this particular task. In addition, all the software components which enabled the vision-based behavior have been detailed. The main contributions of the proposed system are summarized herein: i) A custom UAV has been built with a flexible hardware architecture along with a novel gripper design in order to perform the grasping maneuver. ii) A complete software architecture is designed in conjunction with the hardware design in order to efficient- and autonomously perform a complete mission in the context of SaI applications. iii) A thorough evaluation of the entire system has been carried out in both simulated, and real flight environments, leading to a successful performance in an international robotics competition.

Concretely, the task of autonomous object grasping from a non-cooperative target UAV which is flying at high speeds has been selected as the baseline task. This mission has been extracted from the 2020 Mohammed Bin Zayed International Robotics Competition (MBZIRC)\footnote{\url{https://www.mbzirc.com/challenge/2020}}, in which several members of the Computer Vision and Aerial Robotics (CVAR) group have participated. In particular, the task of autonomous object grasping corresponds to the Challenge 1 (out of three challenges) defined in this competition. Our team ended up in the 3$
^{rd}$ position in the Grand Challenge\footnote{\url{https://www.mbzirc.com/winning-teams/2020/challenge4}}, which is a parallel combination of the proposed three challenges of the competition.

The remainder of the paper is organized as follows: Section \ref{sec:related} provides a literature review, analyzing the most relevant related work. Section \ref{sec:problem_statement} states the problem definition this work aims at solving. Section \ref{sec:hw_setup} describes our proposed hardware platform, detailing all the sensory and actuation resources. Section \ref{sec:system_description} describes the developed software stack and their components. Section \ref{sec:experiments_and_res} outlines
the carried-out experiments and their corresponding results
and Section \ref{sec:discussion} remarks on and discusses the most relevant
experimentation outcomes. Finally, Section \ref{sec:conclusions} concludes the
paper and indicates future lines of research.

\section{Related Work}
\label{sec:related}

The present work has been developed in the context of fully autonomous aerial robotics, with a broad coverage of state-of-the-art techniques. On this basis, several studies in the literature provide methods and techniques which are adjacent to the present study. In the following section, the most relevant related work is included, with a special emphasis on the tangential aspects to the methods provided.

\subsection{High-Speed Navigation for Autonomous UAVs}

In the broad context of robotics, multirotor aerial robots particularly provide a wide maneuverability, being technically able to navigate at high speeds due to the low friction of air and their 6-DOF design. Nevertheless, autonomous navigation at such high speeds in cluttered or interactive environments pose an enormous challenge not yet solved by the research community. In this regard, there are several studies in the literature which are able to achieve notable velocities in fully autonomous mode.

Autonomous navigation in cluttered and/or unstructured environments has been thoroughly studied in the literature. However, high-speed navigation in such environments has been recently explored. In \cite{do2019high}, Parrot Bebop multirotor was able to handle dynamic obstacles in indoor environments, at on average 1 m/s and using an RGB camera as the main sensor. In \cite{liu2016high}, fast and efficient path planning algorithm for UAV navigation in unstructured environments was proposed. It provided a peak velocity of 3 m/s using an RGB-D sensor. Smoothing And Mapping With Inertial State Estimation (SAMWISE) navigation system was proposed in \cite{steiner2017vision}, achieving 20 m/s in an outdoor environment. It was aided by heterogeneous sensory inputs for state estimation, such as LIDAR, RGB-D and Inertial Measurement Units (IMUs). In \cite{florence2020integrated}, an MPC scheme and local estimation were utilized for fast reaction, achieving a peak velocity of 5.5 m/s in simulation.

As multirotor hardware design is potentially able to provide agile or aggressive maneuvers, some studies have researched on the generation of autonomous acrobatic movements. In \cite{kaufmann2020deep}, acrobatics maneuvers were trained in simulation with no expert demonstrations and deployed in an outdoor environment. The ego-motion was provided by feature tracks and the IMU allowed for motion recovery. In \cite{falanga2017aggressive}, on-board localization and perception was utilized in order to fly through tilted narrow gaps. The window estimation was achieved through an RGB camera, and a maximum velocity of approximately 2.6 m/s was reached. In \cite{loianno2016estimation}, high-speed narrow gap crossing was achieved at a maximum speed of 4.5 m/s. Visual-Inertial Odometry (VIO) was used for state estimation and the window was not being actively detected. 

In the 2017 Mohammed Bin Zayed International Robotics Competition (MBZIRC), a fast landing maneuver on top of a moving platform challenge was proposed. Several research groups have tackled the challenge with a variety of techniques. In \cite{beul2017fast}, a DJI Matrice 100 was proposed for landing at a maximum speed of 2 m/s. In \cite{cantelli2017autonomous}, Tracking-Learning-Detection (TLD) was used along with a pre-trained model for detection. The team reached 4$^{th}$ position, having autonomously landed in 140 seconds. In \cite{baca2017autonomous}, computer vision techniques and an MPC scheme are used, having provided a complete landing maneuver in 25 seconds. In \cite{beul2019team}, two RGB cameras were utilized for detection and a maximum speed of 2 m/s was achieved. In the stated reference, the team reached 1$^{st}$ position in the Grand Challenge and 3$^{rd}$ position in two sub-challenges. In \cite{bahnemann2019eth}, VIO and MPC methods were implemented, with a maximum velocity of 8 m/s.

In this work, the main sensor for target state estimation has been an 3-DOF RGB camera gimbal. The complete approach has reached a maximum speed of approximately 6 m/s in outdoor space when following a moving target.

\subsection{Object Grasping for Multirotor Systems}

Object interaction, and concretely object grasping, in the context of multirotor systems poses a notable challenge, since it requires several components to be performing simultaneously, from complex controllers to real-time perception algorithms. In this respect, several studies have provided object grasping solutions for certain scenarios.

Multiple grasping techniques and maneuvers are provided in the literature. In \cite{liaro2019}, a claw-inspired design has been used to perform the grasping of cylindrical objects with UAVs. In \cite{thomas2013avian}, the 2-DOF gripper was inspired on the claw of an eagle. The dynamic model was determined using Lagrange mechanics and the trajectory was computed by restricting positions in space. In \cite{ramon2017detection}, a 3-DOF arm and a RGB-D camera sensor was utilized for 3D estimation of the objects. In \cite{pounds2014aerial,pounds2011yale,pounds2011grasping}, a robotic arm with no DOF was designed for helicopter grasping. PID scheme was used for control. In \cite{parra2013toward}, multiple UAVs were used for object grasping, with advanced non-linear control for swarms in simulation. In \cite{spurny2019cooperative}, an Unscented Kalman Filter (UKF) for object state estimation and prediction was used. MPC control was proposed and RTK for positioning. In \cite{spica2012aerial}, a planning strategy relying on differential flatness was then proposed to concatenate one or more grasping maneuvers by means of spline-based subtrajectories. In \cite{pounds2011practical}, a preliminary analysis and experiments for reliable grasping of unstructured objects with a robot helicopter was presented. Key problems associated with this task were discussed, including hover precision, flight stability in the presence of compliant object contact, and aerodynamic disturbances. In \cite{gioioso2014flying},  a flying hand was presented, where a robotic hand consisting of a swarm of UAVs able to grasp an object. Each UAV contributes to the grasping task with a single contact point at the tooltip. In \cite{kim2013aerial}, a quadrotor with 2-DOF robotic arm was proposed of manipulation. In \cite{bodie2019omnidirectional}, omnidirectional aerial manipulation platform for robust and responsive interaction with unstructured environments, toward the goal of contact-based inspection is proposed. In \cite{yang2014dynamics}, a novel backstepping-like end-effector tracking control law, which allowed to assign different roles for the center-of-mass control and for the internal rotational dynamics control according to task objectives was designed. In \cite{fumagalli2014developing}, an impedance controller for grasping with a 3-DOF arm is described.

In this study, a novel gripper, placed above the quadrotor's center of mass and statically fixed to its frame has been proposed. This design allows for fast grasping maneuvers (from a lower position with respect to the target object) of hanging objects and enables a smooth movement throughout the whole grasping trajectory without the necessity of abrupt position, velocity or accelerations direction changes.

\subsection{Vision-Based Target Following for Autonomous UAVs}

Autonomous target following constitutes an increasing matter of interest due to the successful application of UAV applications in the industry. On this subject, following an unknown or non-cooperative target is a challenging task which involves complex perception and control systems. In the literature, a number of studies have addressed the task of autonomous object following.

Some approaches rely on an online user-defined Region-of-Interest (RoI) within the image plane during the execution of a real flight in order for the UAV to be able to follow a tracked object. PID control scheme was used to control a multirotor aerial robot in \cite{pestana2014computer} based on the OpenTLD tracker in the image plane. In \cite{li2016monocular}, the authors developed a novel frequency and image-plane-domain tracker with latter pose estimation of the object and standard PID control. A state-of-the-art tracker was used in \cite{mueller2016persistent} and a standard PID control scheme within the image plane was established. The stated work also performed real-flight validation tests and developed a novel camera handover strategy in order to enable long-term operation through several UAVs. In \cite{haag2015correlation}, correlation filters for short-term tracking and a re-detection strategy based on TLD was utilized for IBVS control.

Classical control schemes (e.g. PID) along with computer-vision techniques (e.g. color-based detectors) have been also explored in the literature. In \cite{teuliere2011chasing}, a color-based detector fed a particle filter that aided a pose-based PID control scheme. A miniature robotic blimp was used in \cite{yao2017monocular} as the robotic platform, with human-pose estimation and standard PID control. In \cite{nageli2018flycon}, human tracking was carried out over a long period of time and long distances by means of active infrared markers and the estimation of the pose of both human and UAVs without calibrated or stationary cameras. In \cite{mondragon20113d}, an IBVS controller and a color-based detector were developed for object following. In \cite{olivares2011aerial}, the fuzzy servo controller was explored along with a color-based detector. A nonlinear adaptive observer was used in \cite{choi2014uav} for target-object estimation and a guidance law for target-object tracking was developed. In this case, the vision sensor was modeled and the complete approach was validated under numerical simulations.

In this work, a camera gimbal aids a machine learning based algorithm for target detection, among others. MPC scheme was integrated for obstacle-free trajectory following of the moving target.

\subsection{Fully Autonomous Aerial Sensors and On-Board Computation}

Fully autonomous systems require a certain amount of both computational and sensory resources. Nevertheless, depending on the employed techniques, an autonomous UAV can incorporate either heavier extereoceptive and/or lighter proprioceptive sensors for performing the high-level mission. Several strategies have been found in the literature to this respect.

LIDAR sensor provides rich and precise information about the surrounding environment. Due to this fact, LIDAR is a commonly utilized sensor in fully autonomous UAV designs. In \cite{tomic2012toward}, a Hokuyo laser rangefinder UTM-30LX and two onboard computers (1.6 GHz Intel Atom, 1 GB DDR2 and ARM Cortex-A8, 720 MHz, 512 MB RAM) were used for indoor and outdoor urban search and rescue applications. In \cite{sampedro2019fully} a Hokuyo laser rangefinder UTM-30LX and an Intel NUC6i5SYK featuring a 2.9 GHz Intel Core i5-6260U CPU were integrated. In stated work, an additional an Intel Realsense R200 RGB-D camera and Lightware SF10/A altimeter were included to enable a higher versatility of the system for search and rescue applications in indoor environments. An AISKopter continuously rotating 3D laser scanner and an Intel Core i7, 8GB RAM onboard computer were selected in \cite{klingbeil2014towards}. Additionally, a global shutter camera, an stereo camera pair  and ultrasonic sensors were also integrated.

Other techniques use stereo camera pairs as the main sensor for perception in autonomous missions. In \cite{schmid2013stereo}, the stereo images were processed in an Intel Core2Duo SU9300@1.86GH and Spartan 6 LX75 FPGA. The selected onboard computer was an OMAP3530 ARM Cortex A8@720MHz for indoor and outdoor navigation applications. Bebop 2 was utilized in \cite{bavle2018stereo} for navigation in indoor environments, along with a Intel R200 RGB-D camera and a Parrot Slam Dunk stereo camera pair. In \cite{mcguire2017efficient}, a 4g micro stereo camera pair and STM32F4 processor was integrated for obstacle avoidance and reactive navigation. In \cite{perez2018architecture}, Orbbec Astra stereo vision was utilized for navigation in GPS-denied areas.

Monocular strategies for autonomous navigation are challenging tasks, since the spatial information is not directly available in the raw information provided by the sensor. Nevertheless, stated RGB cameras are low-cost and passive sensors which provide high-dimensional information which can be exploited with different approaches. In \cite{loianno2016estimation}, a downward facing VGA camera with 160º field of view, a VGA stereo camera, and a 4K camera were integrated for estimation, control and planning for aggressive flight (running on a Qualcomm Snapdragon board). In \cite{ross2013learning}, an offboard monocular approach for reactive navigation in cluttered natural environments with an ArDrone 2.0 was proposed. In \cite{von2017monocular}, an offboard monocular SLAM based technique was implemented for navigation in cluttered indoor environments with an ArDrone 2.0.

In this study, the main sensors for ego-motion and target state estimation have been an RGB camera gimbal, an IMU, an RTK positioning device and two VGA stereo pairs. The selected onboard computer has been an NVIDIA Jetson AGX Xavier. The complete strategy is primarily based on an RGB camera gimbal for searching, detection, target state estimation and grasping maneuver.

\section{Problem Statement}
\label{sec:problem_statement}

\begin{figure}[h]
	\centering
	\includegraphics[width=0.3\textwidth]{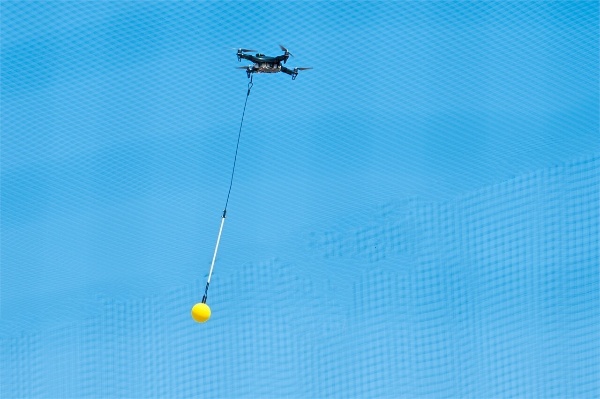}
	\caption{Target UAV with tethered ball in the real competition environment.}
	\label{fig:target_uav}
\end{figure}

\begin{figure}[h]
	\centering
	\includegraphics[width=0.6\textwidth]{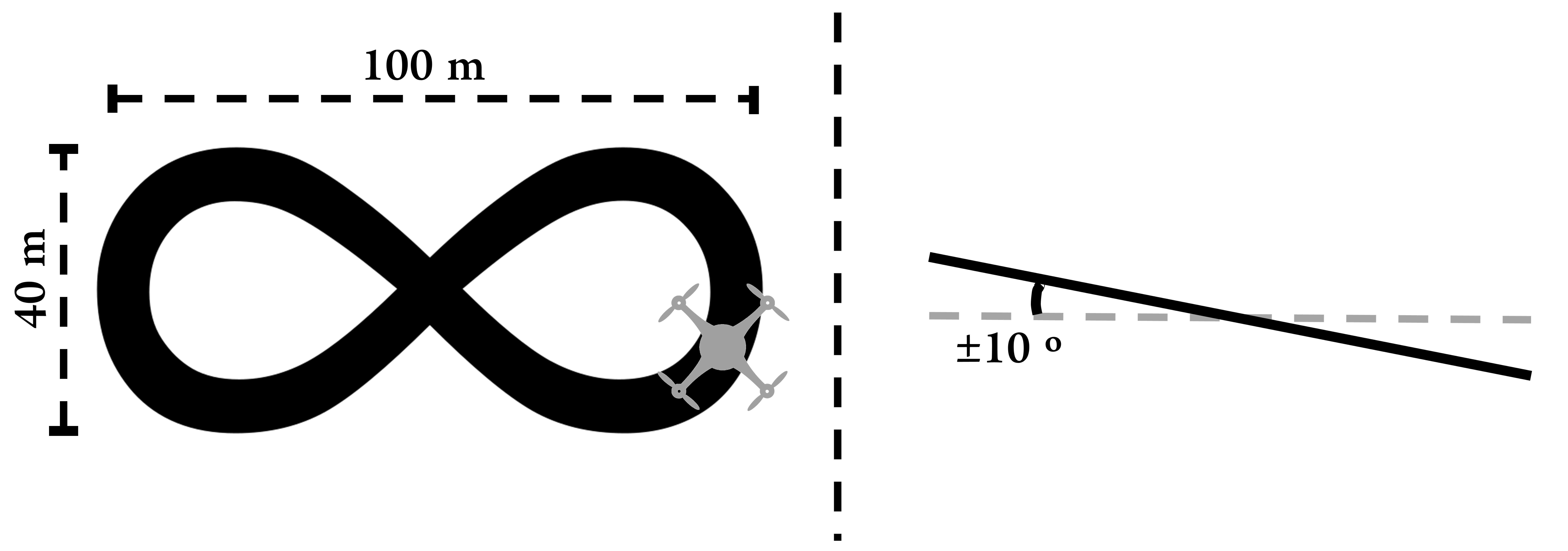}
	\caption{Structure of the designed competition arena for the Challenge 1 in the 2020 MBZIRC competition.}
	\label{fig:arena}
\end{figure}

In this work, the task of autonomous object grasping has been selected as the baseline application. The stated task has been extracted from the Challenge 1, proposed in the 2020 MBZIRC competition. It consists of the grasping of an object which is mounted on a target UAV. The target UAV is flying within a volume of 100 m $\times$ 40 m $\times$ 20 m, describing a lemniscate shape inside a tilted 3D plane, as shown in Fig. \ref{fig:arena}. In the competition, the whole Challenge 1 is designed to last 15 minutes per trial, and the velocity of the target UAV is scheduled to 8 m/s for the first 8 minutes and 5 m/s for the next 7 minutes. In Fig. \ref{fig:target_uav}, the target UAV with a tethered ball in the real competition  environment has been depicted.

The overall objective of the challenge was to autonomously perform a series of interaction tasks using aerial vehicles at high-speeds. The object to be grasped corresponded to a ball of an approximate radius of 11 cm, which is hanging from a cable of approximately 1.5 m in length. Once the item has been removed from the tether, it must then be delivered to a pre-specified 10 m $\times$ 10 m landing location inside the arena. 

Furthermore, the UAV (or a team of maximum three UAVs) will have to locate and physically interact with a set of static objects (balloons of 45 cm in diameter), which are of unknown color, randomly distributed within the outdoor flight arena, and tethered to the ground. For added difficulty, once these static items are detected, the UAV will have to navigate and burst each balloon using any component installed on-board.

\section{Hardware Configuration}
\label{sec:hw_setup}
\subsection{Aerial Vehicle Platform}

\begin{figure}[t!]
    \centering
    \includegraphics[width=0.75\textwidth]{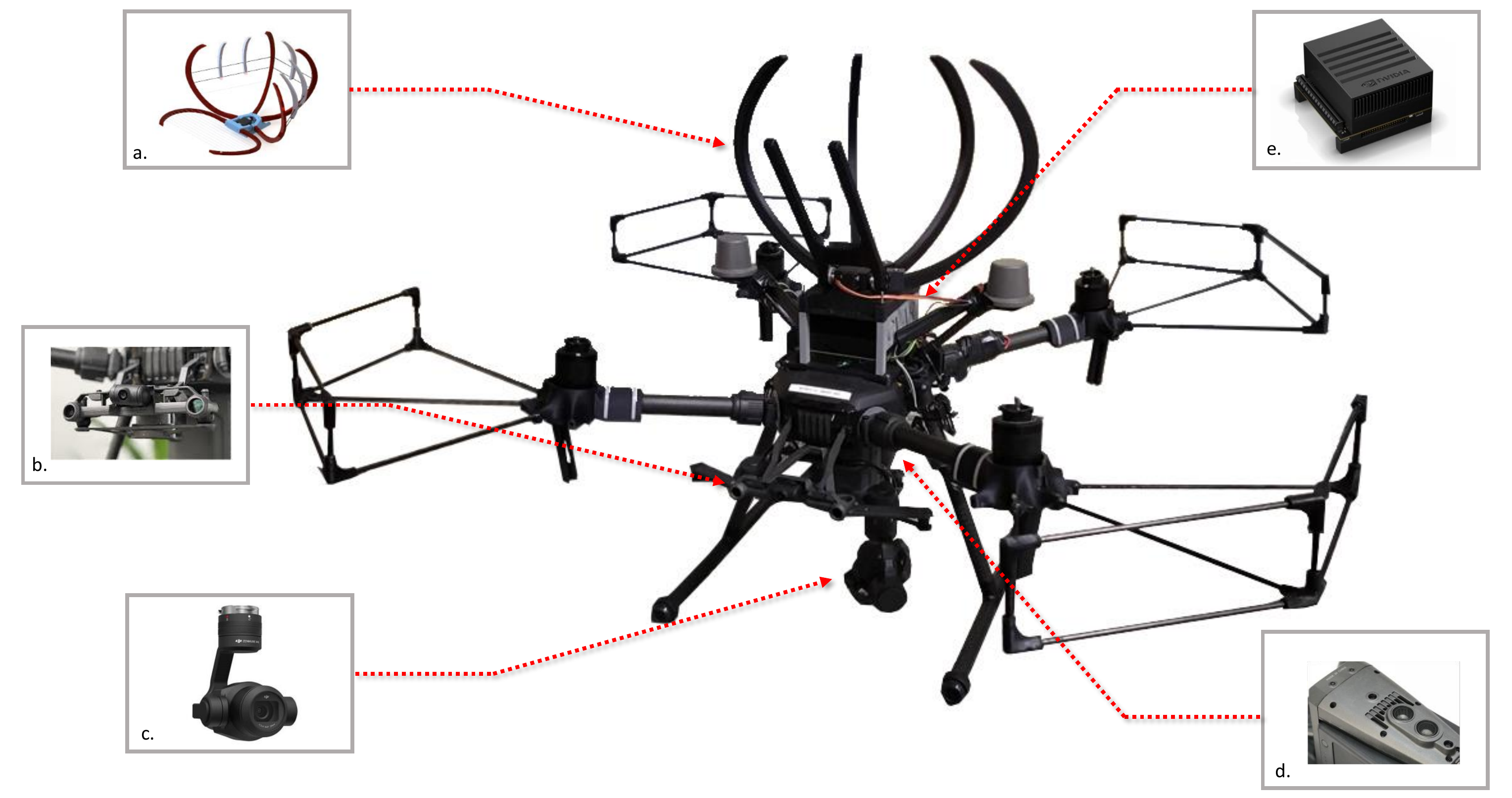}
    \caption{Aerial platform used for Challenge 1 with the on-board components: (a) Gripper with a Servo Mechanism (b) DJI RGB Stereo Pair (c) DJI X4s RGB Camera (d) Sonar Altimeter \& Bottom Camera (e) Nvidia Xavier.}
    \label{fig:aerial_platform} 
\end{figure}

The aerial platform used for the Challenge 1 of the MBZIRC competition was a modified version of the DJI Matrice 210 RTK v2 UAV, shown in Fig.~\ref{fig:aerial_platform}.~This robust platform was chosen as the main aerial vehicle due to the demanding flight performance required by the different tasks, \textit{i.e.}, flights up to  10 m/s, large payload, and extended flight times. The aerial robot was equipped with RTK and GPS receivers, DJI Visual Position System (VPS), sonar altimeter and a high resolution Zenmuse X4s on-board camera featuring a 20MP, 1" sensor and 84 degree FOV high-resolution lens. 

Additionally, the aerial platform was equipped with the Single Board Computer (SBC) NVIDIA Jetson AGX Xavier with an 8-core ARM v8.2, 64-bit CPU running Ubuntu Linux 18.04 Bionic Beaver for on-board computing and a 512-Core Volta GPU for image processing tasks. All inter-process communications were handled using the Melodic Morenia distribution of the Robotic Operating System (ROS) middleware. 

As previously mentioned, the objective of Challenge 1 was to retrieve an item tethered to a separate target UAV in motion. To this effect, the DJI Matrice 210 UAV was equipped with a custom designed gripping mechanism, shown in Fig. \ref{fig:aerial_platform}, which enabled the grasping of the item. It was also equipped with custom designed protections in order to protect the propellers of the aerial robot against collisions from the target UAV or the tethered item.~Section.~\ref{gripper_design} and Section.~\ref{protections_design} explain in detail the design mechanisms for the gripper and the protections respectively. 


\subsection{Gripper Design}
\label{gripper_design}
\begin{figure}[htb!]
	\centering
	\subfigure[]{\includegraphics[height = 5cm, width=0.3\textwidth]{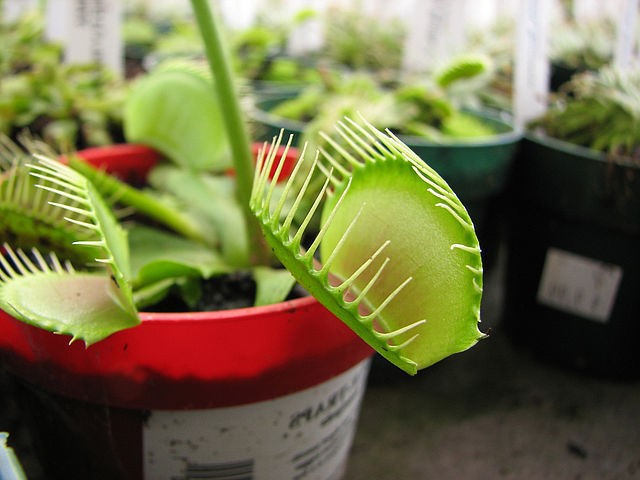}\label{fig:VenusFlytrap}}
	\hspace{1cm}
	\subfigure[]{\includegraphics[height = 5cm,
	width=0.40\textwidth]{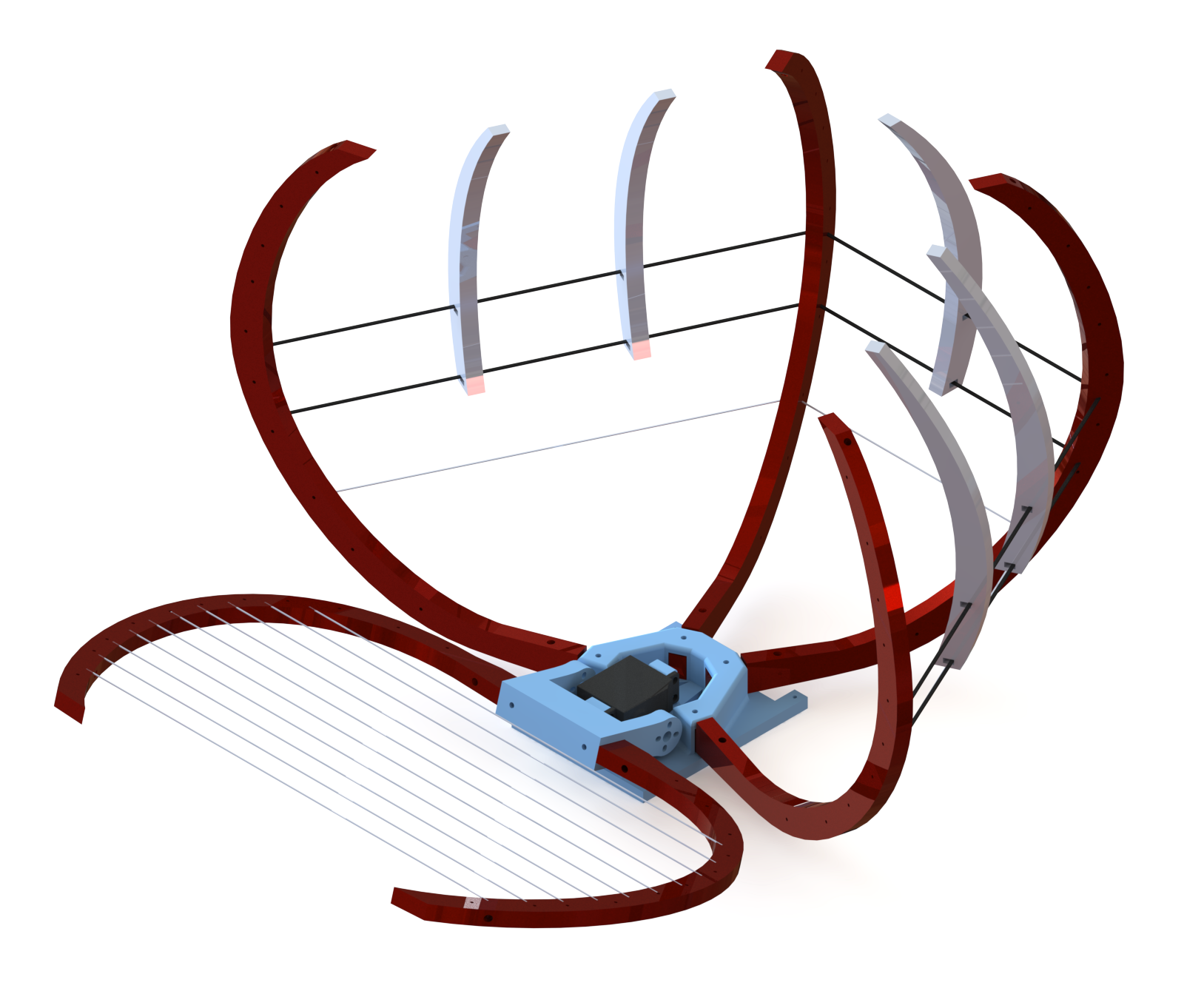}\label{fig:GripperCAD}}
	\caption{(a) Venus Flytrap. (b) Gripper CAD}
	\label{fig:gripper_cad}
\end{figure}

A bio-inspired design has been selected for the gripper actuator. There exist carnivorous plant species, such as the Venus Flytrap (Fig \ref{fig:VenusFlytrap}), which are able to trap flying insects without the necessity of performing fast grasping movements. The stated species have two moving parts which are able to be fully closed, trapping the insect which was flying in between. In order to avoid the insects from escaping, there are sharp and interlocking ends in borders of each part. 

The gripper design is required to provide a wide volume for catching the ball with the minimum weight possible. Also, it is required to allow the tether to pass through the gripper, in order to successfully release the ball from the target drone and trap it into the gripper. The gripper is composed by two main parts, one mobile and one static. On the one hand, the cage, which is the static part, consists on 4 long \textit{fingers}  followed by 5 small \textit{fingers} bounded together with carbon fiber bars and epoxy resin. Moreover, nylon threads were used for covering the bottom part of the cage, leaving free space between the upper part of the long fingers. This free space between fingers allows the ball tether to flow and go through the cage. On the other hand, the mobile part of the gripper consists on one planar surface built with 2 \textit{fingers} joined with nylon. When the gripper is closing, this planar surface rotates into the cage, catching the ball inside it. A servomotor was used for closing the gripper. Stated servo is controlled trough an Arduino Nano, in charge of closing and opening the griper whenever the high-level mission components require to catching maneuver. At last, a single point laser sensor pointing upwards was placed inside the gripper, in order to detect if the ball is trapped inside the gripper. 
\begin{figure}[htb!]
    \centering
    \includegraphics[width=0.7\textwidth]{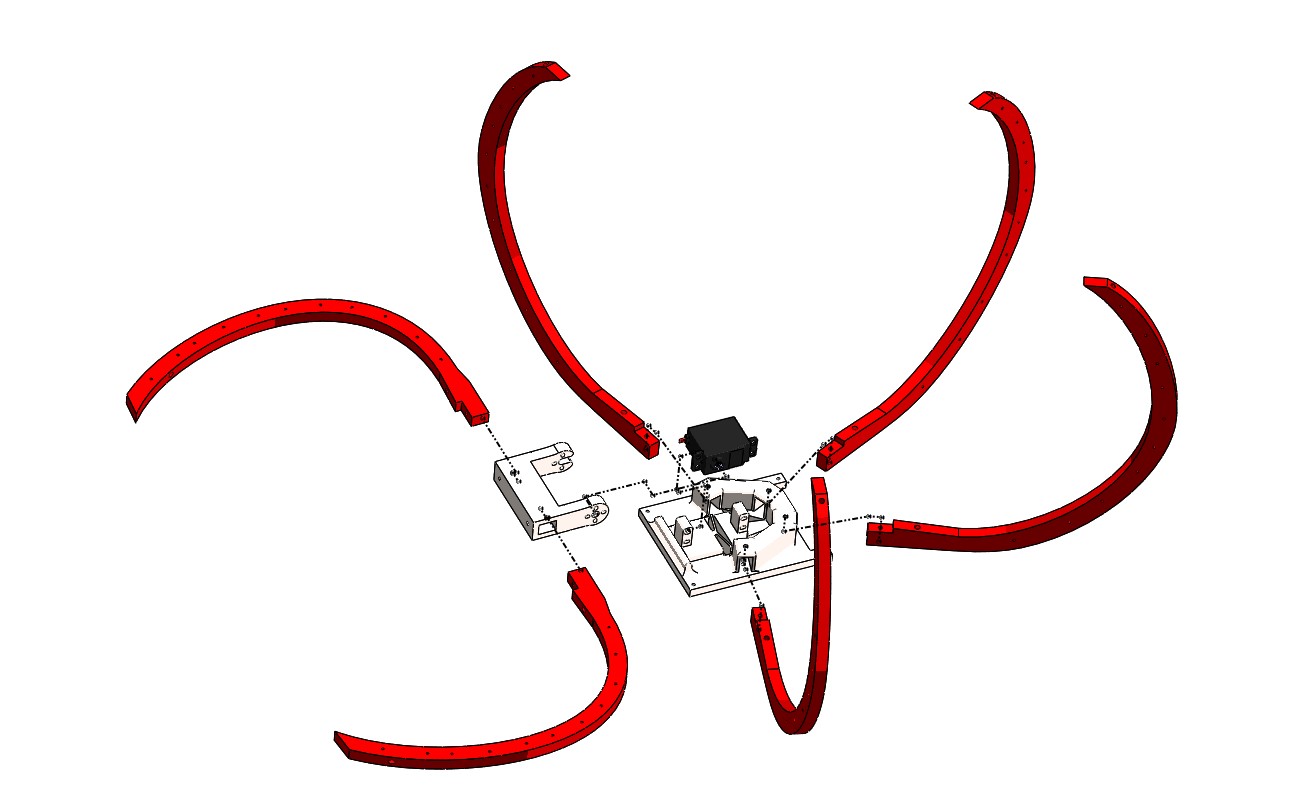}
    \caption{Exploded view of the main structural components of the gripper.}
    \label{fig:exploded_view_gripper} 
\end{figure}

PolyLactic Acid (PLA) 3D-printed pieces for the whole design of the gripper have been utilized, except for the carbon fiber bars and the nylon wires. PLA was used, over other 3D printing materials, due to its ease of handling and its rigidness. Carbon fiber bars were used because of its lightweight and tenacity, and were attached to the \textit{fingers} using epoxy resin. In order to cover open patches (to avoid the ball dropping out of the gripper), nylon threads were used. Nylon can resist the impacts without stretching or breaking.

\begin{figure}[htb!]
    \centering
    \includegraphics[width=0.35\textwidth]{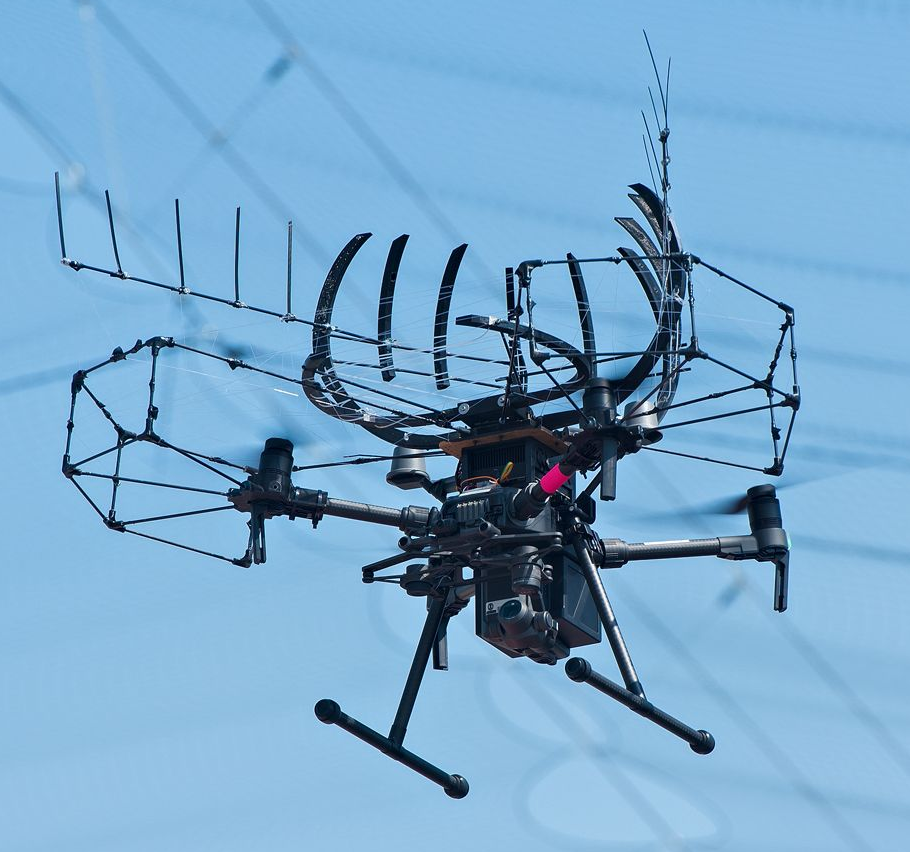}
    \caption{The complete proposed platform, with the gripper and all protections included in the real competition environment.}
    \label{fig:aerial_platform_flying} 
\end{figure}

\subsection{Protections Design}
\label{protections_design}

As explained in Section \ref{sec:problem_statement}, the challenge was composed of one main task and another complementary task. However, both tasks required physical interaction between the aerial vehicle and other objects, such as balls or balloons. This physical interaction implies the risk of propellers touching hard objects, such as the pole which sustained the balloons or the bar which attached the ball to the moving UAV. These structures may cause a propeller to break, which could lead to a UAV full failure or crash.

The risk of failure was a strong motivation for the design of custom protections for both aerial platforms. Stated protections have to accomplish design constraints. First, the propellers must be protected from physically touching any pole, bar or ball. Second, they must be lightweight and easy to assemble, disassemble and repair. Two configurations for the protections design were built, which are related to the task to be carried out, and are as follows:

\begin{figure}[htb!]
	\centering
	\subfigure[]{\includegraphics[height = 5cm, width=0.40\textwidth]{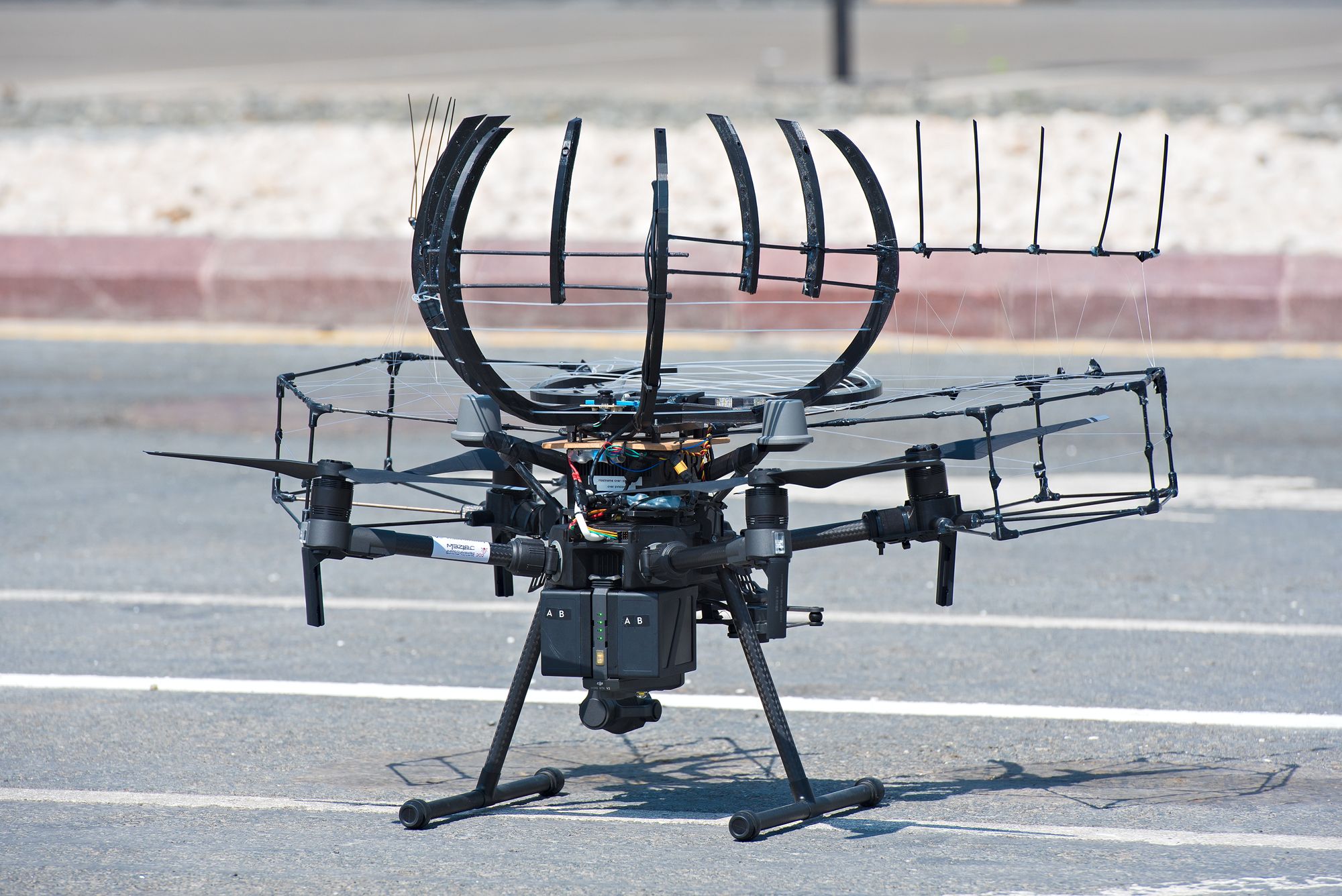}\label{fig:Full_covered}}
	\hspace{1cm}
	\subfigure[]{\includegraphics[height = 5cm,
	width=0.40\textwidth]{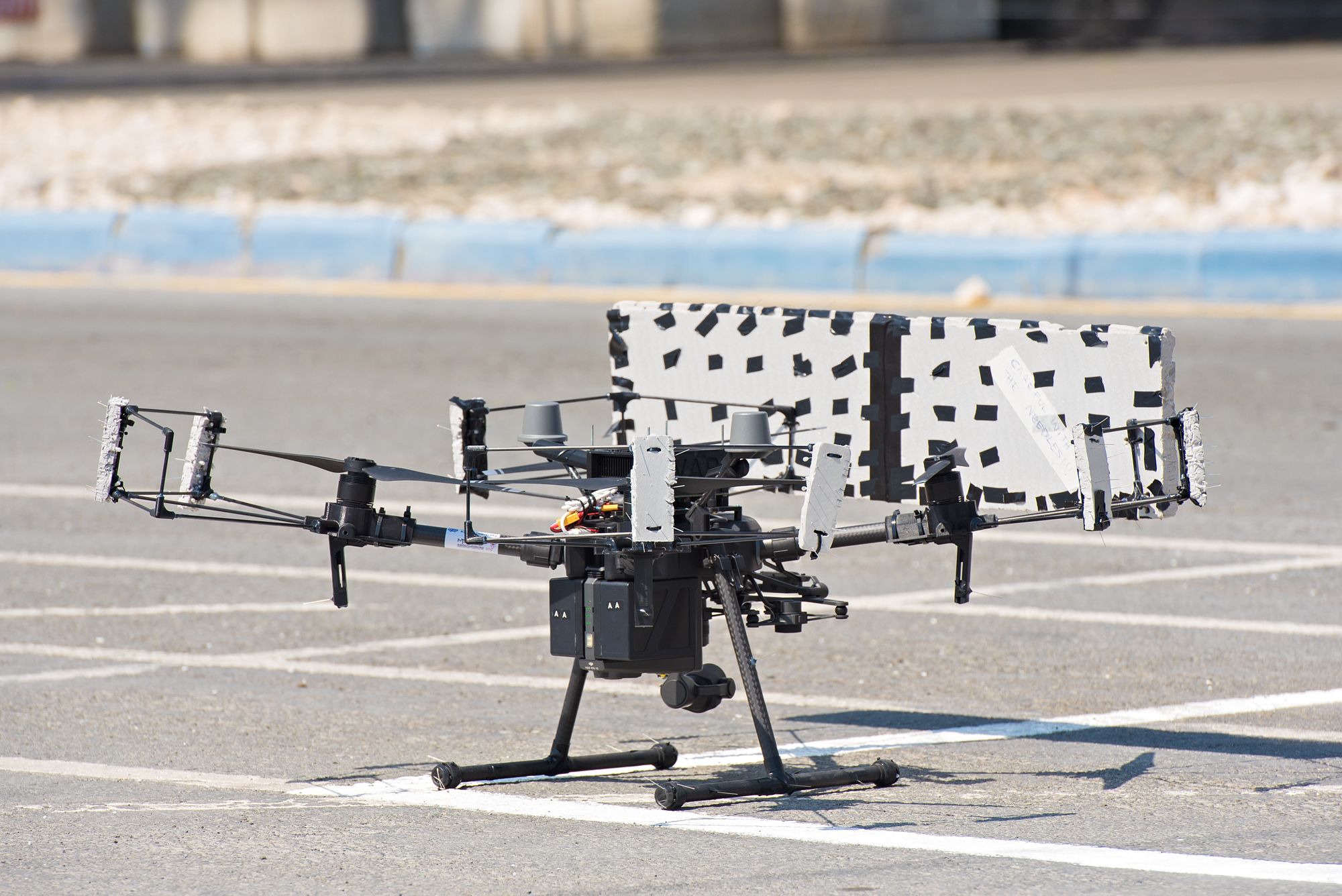}\label{fig:cover_protections}}
	\caption{(a) Full covering front protections. (b) Contact covering protections with needle plates}
	\label{fig:protections}
\end{figure}

\begin{itemize}
    \item \textbf{Full covering}: This configuration is meant for the ball grasping task, where the UAV is required to intercept a target UAV with a ball attached to it. In stated task, the protections fully cover the UAV arm, including the propellers, avoiding them to break by touching other objects while they are rotating at high speeds. These protections are attached to the UAV arms, and to a plate placed between the on-board computer and the gripper. Attaching the structure to these points improves the rigidness of the protections, preventing them from bending and colliding with the propellers. Additionally, all the gaps between the carbon fiber bars were covered with nylon, avoiding the risk of any object impacting with the propellers (Fig. \ref{fig:Full_covered}).
    
    \item \textbf{Contact covering}: For the objective of bursting balloons, the main threat are the poles which support stated balloons. This poles were static on the ground, which reduced the risk of any object colliding with the tops of the propellers. In this case, lighter protections, which were easier to replace when needed, were selected. In this case, four isolated protections, one attached to each UAV arm, covered the propellers in case of the robot collisions with the poles (Fig. \ref{fig:contact_cover_protection_cad}).

    \begin{figure}[htb!]
        \centering
        \includegraphics[width=0.4\textwidth]{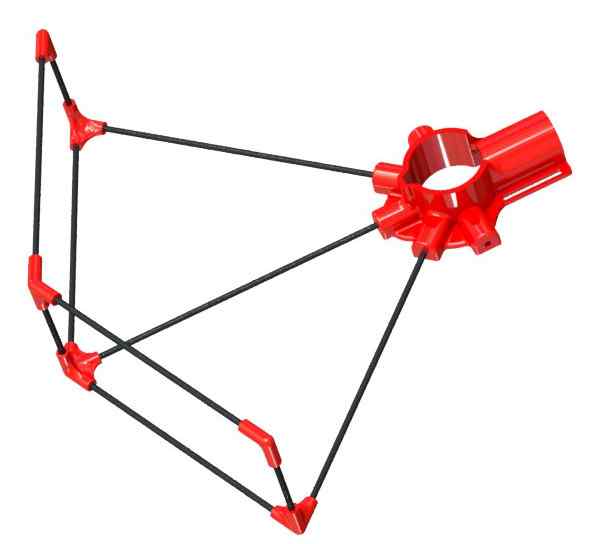}
        \caption{CAD of one isolated contact covering protection}
        \label{fig:contact_cover_protection_cad} 
    \end{figure}

    In order to pierce the balloons, multiple extruded polystyrene plates with several needles were attached along all the protections, with one wide plate between the front protections in order to burst the balloons by frontally colliding with them  (Fig. \ref{fig:contact_cover_protection_cad}).

\end{itemize}

    
    


\section{System Description}
\label{sec:system_description}
In order to perform the complex tasks encountered in Challenge 1 of the MBZIRC, a robust and flexible system architecture is essential for efficient development and validation of both software modules and hardware interfaces.  The system architecture that is presented in this work has been developed using the Aerostack framework \cite{ref_aerostack_1}\cite{ref_aerostack_2}. 

Aerostack is an open-source multi-purpose software framework for the development of autonomous multi-robot unmanned aerial systems created by the Computer Vision and Aerial Robotics (CVAR) group. The Aerostack modules are mainly implemented in C++ and Python languages and are based on Robot Operating System (ROS) for inter-communication between the different components. To provide higher degrees of autonomy, Aerostack’s software framework integrates state-of-the-art concepts of robotics and artificial intelligence. Aerostack has been used in the development of complex robotic systems related, for example, to natural user interfaces \cite{ref_natural_user_interfaces}, surface inspection \cite{ref_surface_inspection}, coordinated multi-robot systems \cite{ref_dynamic_mission_planning}, landing on moving platforms \cite{ref_landing_moving_platform}, and search-and-rescue missions \cite{ref_search_and_rescue}. For a deeper understanding of Aerostack, we refer the reader to the extensive documentation and publications that are  available on its web site\footnote{\url{www.aerostack.org}}. It has to be noted that, although Aerostack framework is able to provide predefined components and intercommunication methods to provide autonomy to UAVs, the components that are described in this work were completely designed and developed for the aim of this competition. All the components described in this study pose a novel contribution of this work.

\begin{figure}[htb!]
\begin{center}
\includegraphics[width=0.90\textwidth]{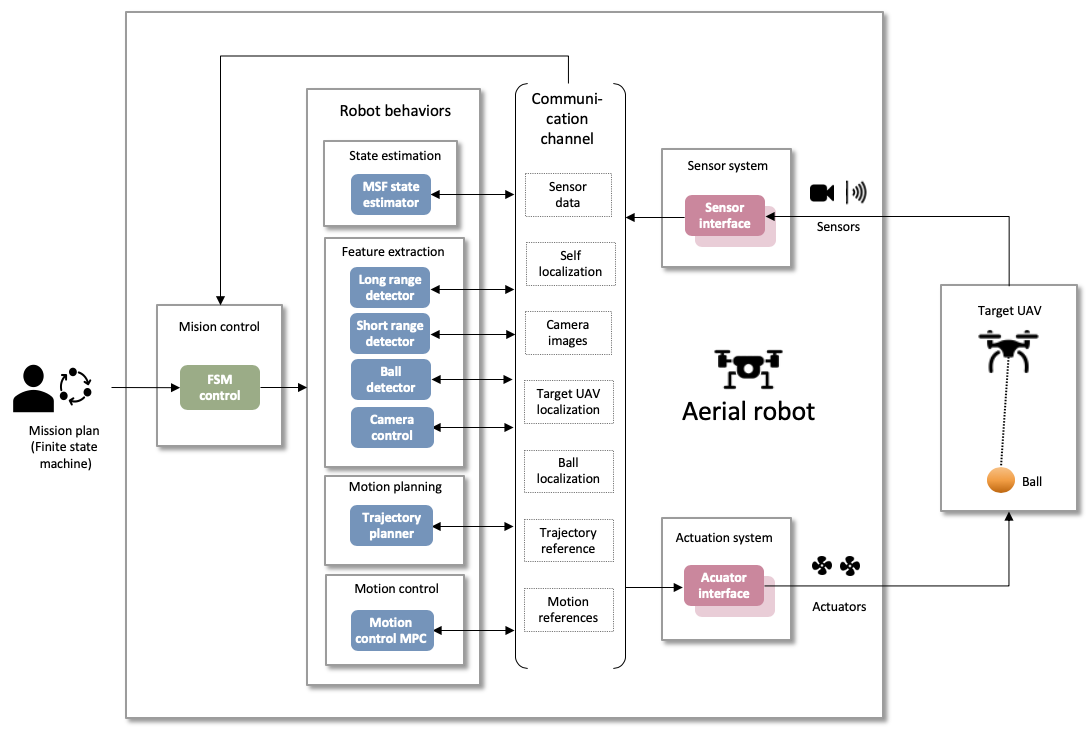}
\caption{System architecture design used for Challenge 1 (based on the Aerostack software framework).}
\label{fig:system_architecture}
\end{center}
\end{figure}

Using the Aerostack software framework, a new system architecture design was developed for the tasks presented in the challenge. Fig. \ref{fig:system_architecture} shows the functionalities that have been implemented in this work. In the figure, colored rectangular boxes represent data processing units (or processes in short) that are implemented as ROS nodes. They are organized in the following main components:

\begin{itemize}
\item \textit{Sensor-Actuator Interfaces:} to receive data from sensors from the aerial platform and send commands to robot actuators. 
\item \textit{Communication Channel:} Based on the Aerostack framework, our architecture uses a common communication channel that contains shared dynamic information between processes. This channel facilitates process interoperability and helps to reuse components across different types of aerial platforms. The channel is implemented with a set of ROS topics and ROS messages.
\item \textit{Robot Behaviors:} Robot behaviors implement the robot functional abilities including state estimation, feature extraction, motion control and motion planning.
\item \textit{Mission Control:} Mission control executes a mission plan specified in a formal description. In this implementation, we use a finite state machine. 
\end{itemize}

Next sections describe in more detail the components related to feature extraction, state estimation, motion planning, motion system and mission control.

\subsection{Feature Extraction System}

A critical element required to achieve the proposed tasks is the capability of robustly detecting the target UAV and objects, as well as accurately estimating their location with respect to the aerial platform. The Feature Extraction System (FES) of the system architecture incorporates all of the perception software modules, which are based on computer vision and deep learning techniques. The FES can be divided into five base modules, which are used during different stages of the challenge: (i) the Long-Range UAV Detector, (ii) the Short-Range Detector, (iii) the Target Position Estimation, (iv) the Camera Gimbal Controller, and (v) the Interleaved Execution Policy.




\subsubsection{Long-Range UAV Detector}
\label{sec:long_range}

\begin{figure}[h!]
    \centering
    \includegraphics[width=0.7\textwidth]{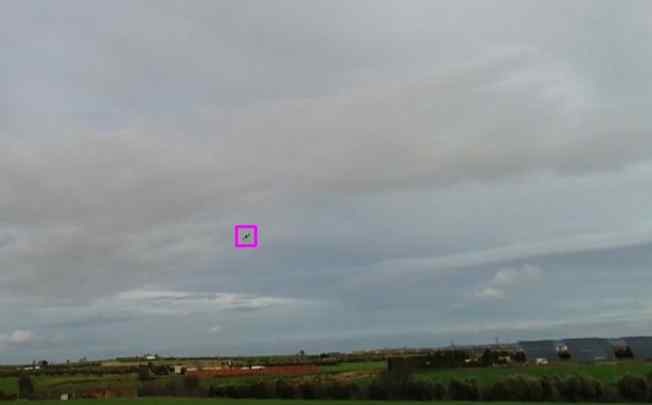}
    \caption{Example of a long-range UAV successful detection. The target UAV appears reduced within the image plane}
    \label{fig:far_detector_inicial} 
\end{figure}

As mentioned in Section \ref{sec:problem_statement}, the area in which the challenge was carried out had dimensions of 100 m $\times$ 40 m. Therefore, in order to detect the target UAV in motion, with such a wide scope area, a long-range UAV detector module has been implemented. The long-range detector is in charge of the detection of moving objects when they appear small in the image plane, e.g., the case of a small UAV flying in a far distance from the camera (see Fig. \ref{fig:far_detector_inicial}). 

\begin{figure}[h!]
    \centering
    \includegraphics[width=0.7\textwidth]{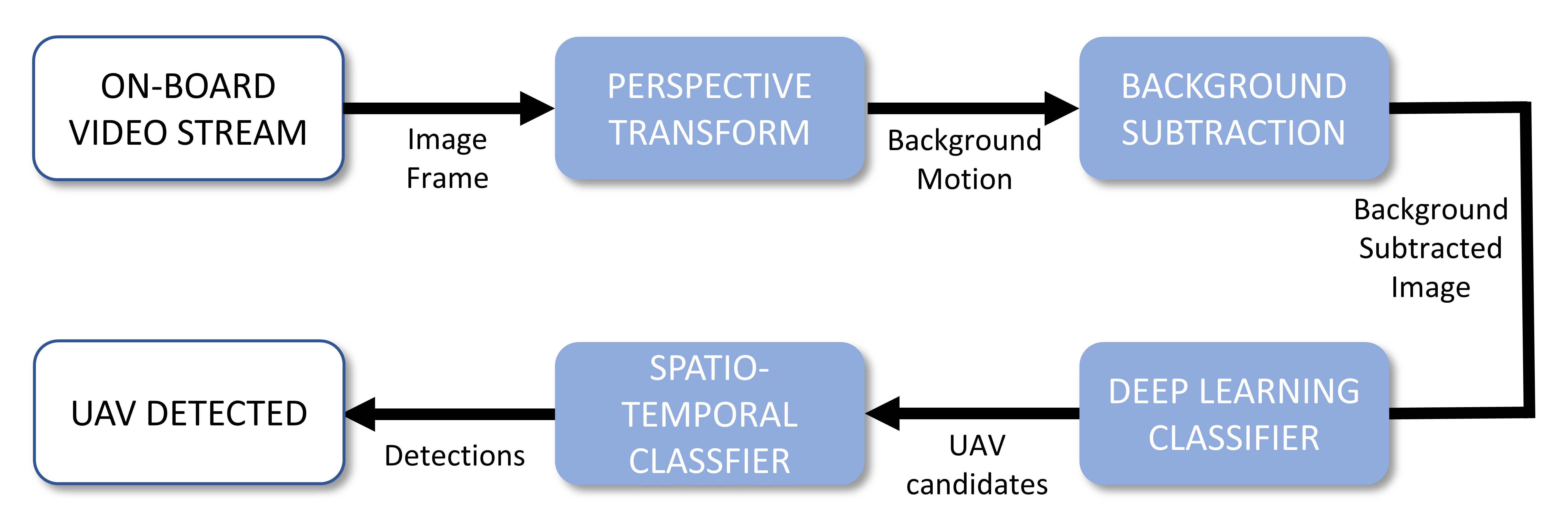}
    \caption{Diagram of the proposed method for the Far UAV detector in Challenge 1.}
    \label{fig:far_detector} 
\end{figure}

In these type of scenarios, where there are no semantics that can be exploited due to small appearance of the object within the image plane, other techniques such as background motion estimation and moving object detection can be incorporated \cite{ref_far_detector}. The processing pipeline is built on 4 steps, which are summarized as follows:

\begin{enumerate}
  \item \textit{Perspective Transform:} The image frames acquired from the on-board camera are used to estimate the background motion between frames. Considering a globally smooth motion with camera projection, the background motion can be calculated using a perspective transform model. For further details of the perspective transform model, refer to  \cite{ref_far_detector}.
  \item \textit{Background Subtraction:} Since the background and the moving objects in the image can be assumed to have fairly different motion models, the moving objects in a far plane can be obtained by compensating the motion of the background. In this regard, with the pixel-wise background estimation model, the background can be subtracted from the original motion image.
  \item \textit{Deep Learning Classifier:} After subtracting the background from the original image, a deep learning classifier can be used to detect the candidates for moving objects in the highlighted patches of the background subtracted image.
  \item \textit{Spatio-Temporal Classifier:} Finally, to find the local motion of the moving objects, a Lucas-Kanade optical flow algorithm has been applied to the classified candidates from previous step, and hand-engineered spatio-temporal features can be extracted to identify actual UAVs in the image.
\end{enumerate}

The architecture of the deep learning classifier is a standard Convolutional Neural Network (CNN) with 2 convolutional layers and 5 fully-connected layers. The last fully-connected layer has 1 linear unit in order to provide a classification of the target as UAV or background.

 \subsubsection{Short-Range Detector}
 \label{sec:short_range}
 
The Long-Range Detector presented in the previous section has been used to find and approach the target UAV during the challenge. Nevertheless, the techniques used to detect objects in a far plane are not suitable for shorter distances. For object detection on close behavior (less than 15-20 meters) a modified version of Tiny YOLO v3 \cite{ref_yolo} has been used. Tiny YOLO v3 is an extremely fast fully convolutional one-shot object detector, which can process full images (at 720p resolution) to output a set of bounding boxes corresponding to the objects in the image along with a confidence value and corresponding class. The modification consist on changing the stride of the up-sample layer from 4 to 2,
 increasing the resolution of the feature maps in order to improve the detection of small objects, and as a consequence, increasing the detection range. 
Since Tiny YOLO v3 can be used as a multi-class detector, the same detector has been used to identify the target UAV, as well as objects of interest (UAV, ball and balloons).

\subsubsection{Target Position estimation}

The target object 3-DOF position estimation (\textit{i.e.} target UAV, ball or balloon) is performed using a basic trigonometric calculation to estimate an approximate 3D position relative to the aerial vehicle.We define an  ellipsoid around the target UAV of width and height $r_w$ and $r_h$ respectively.
We assume that the projection of the ellipsoid in the image plane corresponds with the bounding boxes provided by the detector (Figure \ref{fig:trigonometric1}) . Letting $x_1$ and $x_2$ be the bottom and top coordinates of the bounding box, we can calculate the angles $\alpha_1$ and $\alpha_2$ (Figure \ref{fig:trigonometric2}, Equation \ref{Eq:single_angle}). Then we can estimate the distance $d$ using this height an the angular coordinate $\alpha_x$. After that, we repeat this procedure using the witdh of the ellipsoid and bounding box, and average both values in order to obtain the final 3D estimation $(d,\alpha_x,\alpha_y)$ . 

\begin{equation}
F=\frac{FOV}{2}
\end{equation}

\begin{equation}
R=\frac{resolution}{2}
\end{equation}

\begin{equation}
f=\frac{R}{tan(F)}
\end{equation}

\begin{equation}
\gamma = tan^{-1}( \frac{x \times tan(F)}{R})
\label{Eq:single_angle}
\end{equation}

\begin{equation}
\beta=\alpha_{2} - \alpha_{1}
\end{equation}

\begin{equation}
\alpha_{x}=\frac{ \alpha_{1} + \alpha_{2}}{2}
\end{equation}
\begin{equation}
d=\frac{r}{tan(\frac{\beta}{2})}
\end{equation}

\begin{figure}[t!]
	\centering
    \subfigure[]{\includegraphics[width=0.3\textwidth]{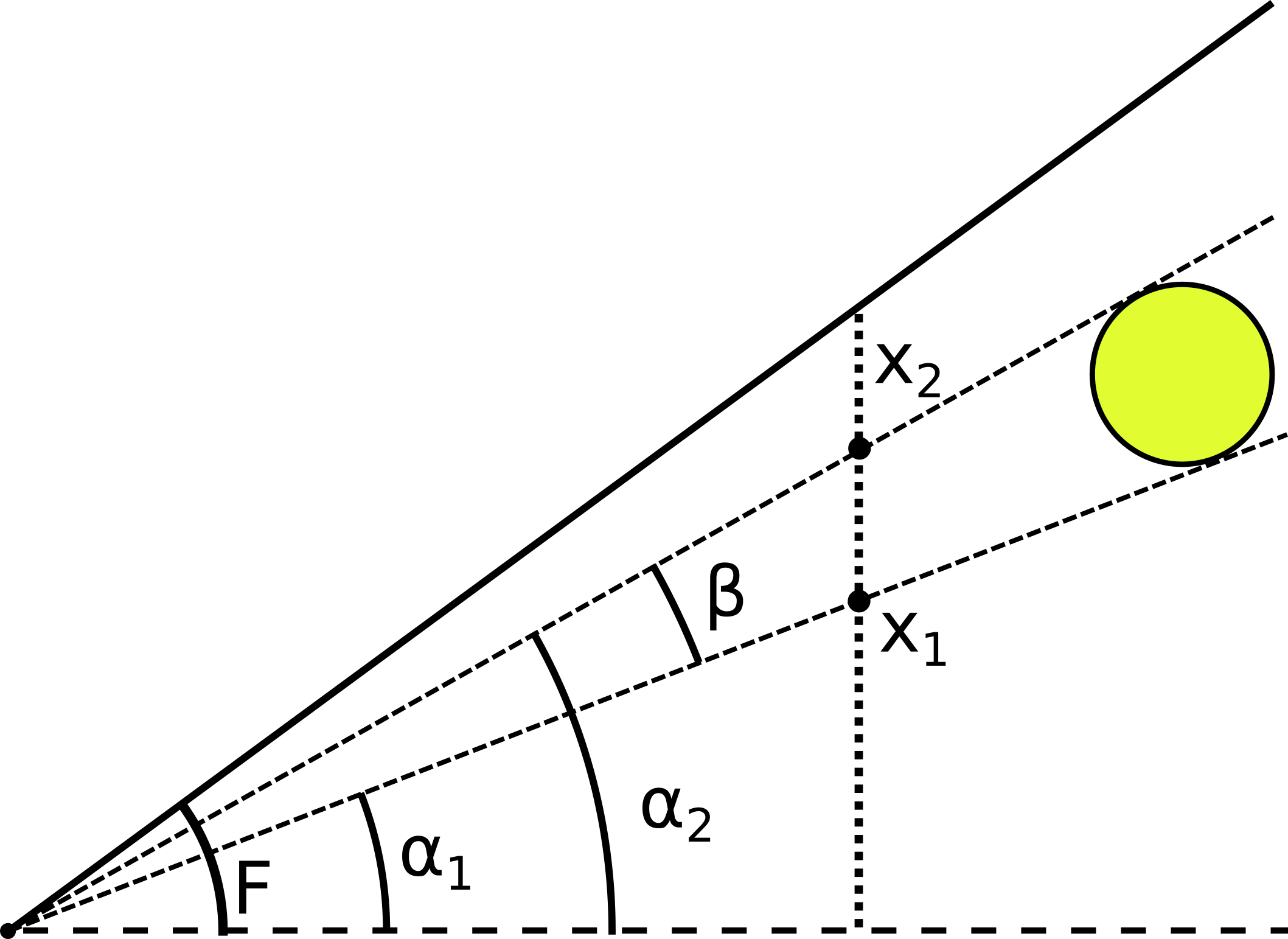}\label{fig:trigonometric2}}
    \hfill
	\subfigure[]{\includegraphics[width=0.3\textwidth]{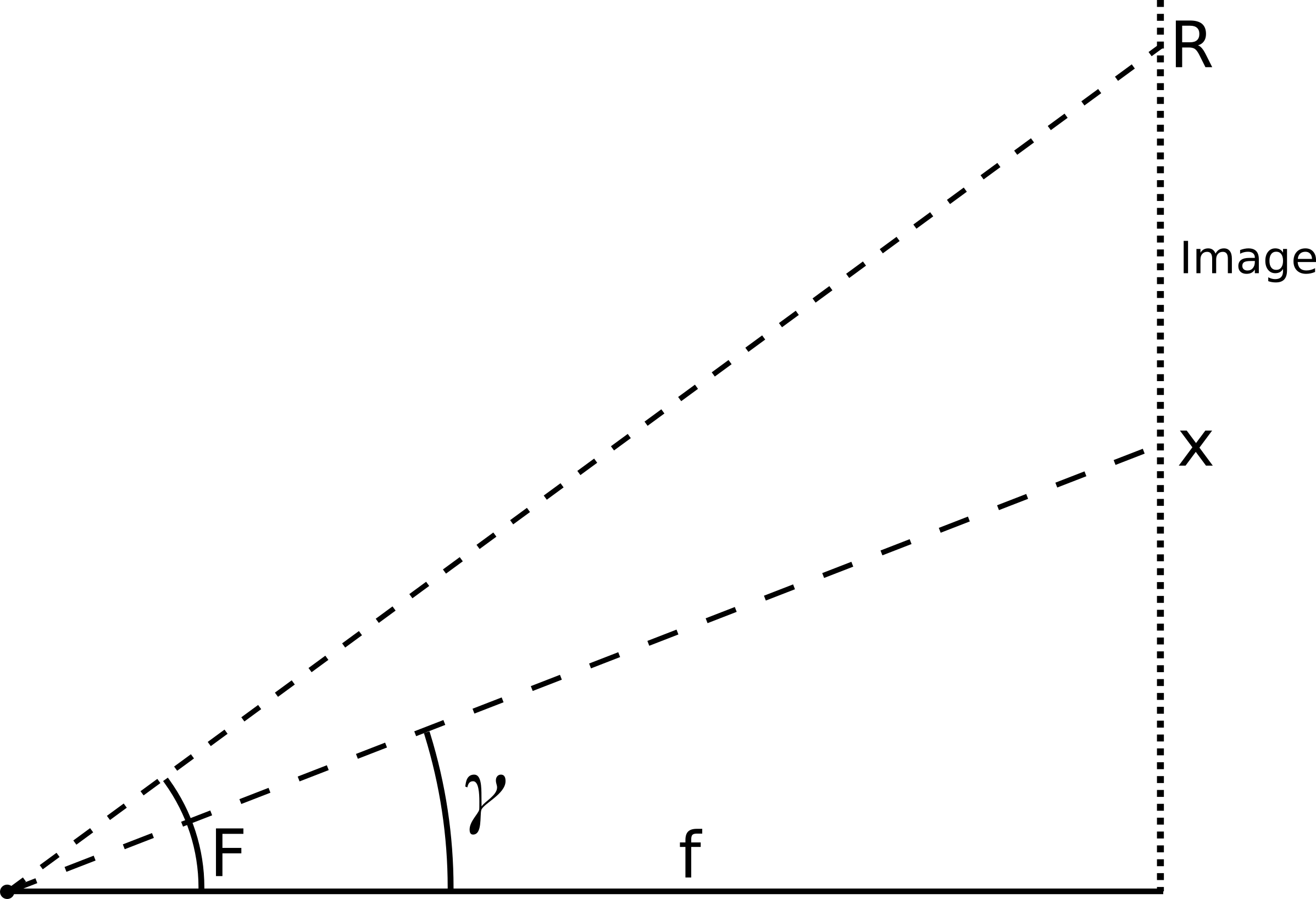}\label{fig:trigonometric2}}
	\hfill
	\subfigure[]{\includegraphics[width=0.3\textwidth]{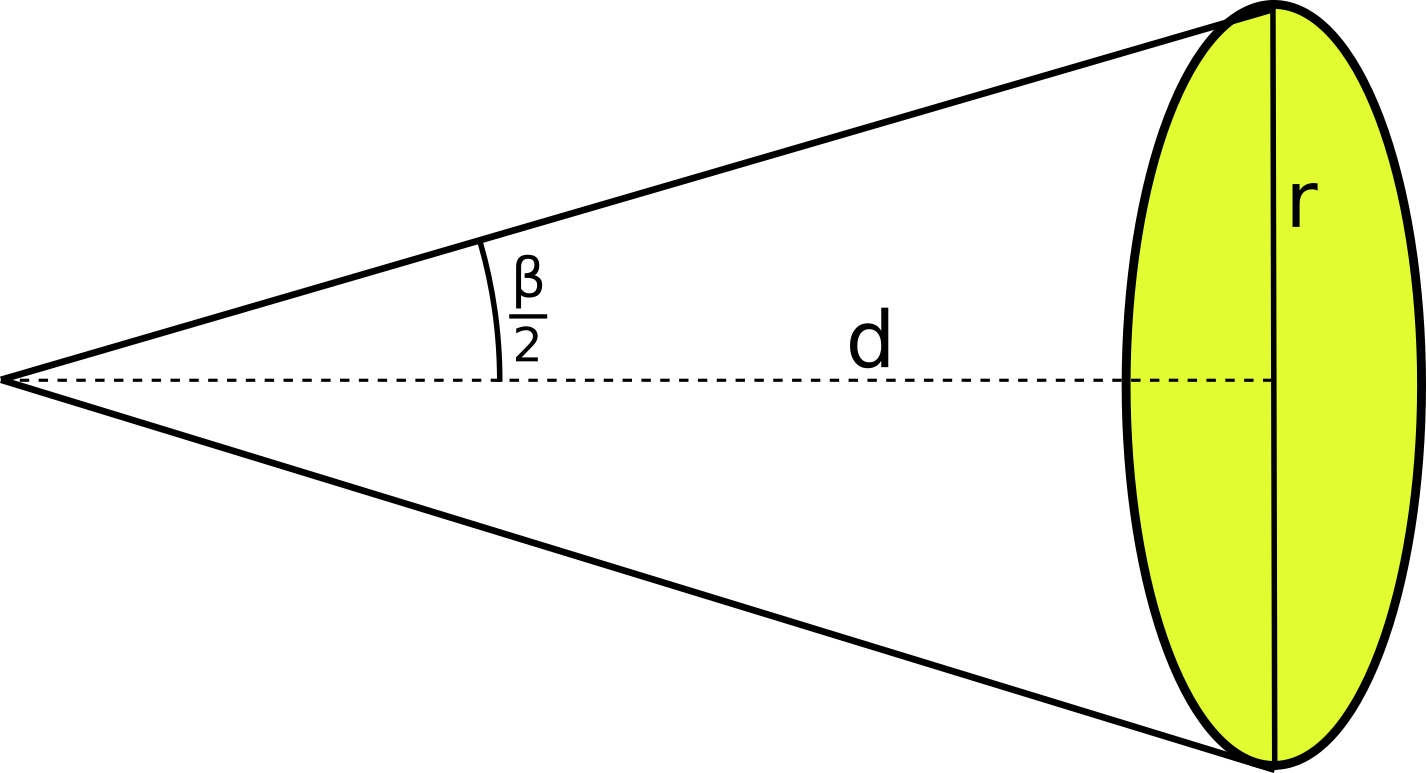}\label{fig:trigonometric2}}
\caption{Diagram of the proposed method object position estimation. (a) An ellipsoid surrounding the object is considered, and the resulting position is calculated using the bounding box of the detection in image coordinates. The vertical line depicts the image plane. (b) Close up diagram of image references.  (c) Close up diagram of distance estimation.}
	\label{fig:planner_sim}
\end{figure}


For the case of the Short-Range Detector (see Section \ref{sec:short_range}), stated trigonometric strategy was complemented by Efficient Perspective-n-Point (EPnP) algorithm\footnote{https://github.com/cvlab-epfl/EPnP} [2], in order to reduce the variance of the estimation, given the high variability found in the bounding box detection within the image plane. On this basis, the four corners of the bounding box in image are paired with the respective points in world (centering the axes at the center of the item). Afterwards the plane which dissects the item (parallel to the image plane) is used, and the resulting 3D corners of a rectangle of the size of the object are matched to the detected bounding box 2D corners to acquire the translation to the camera frame of reference. Finally, a linear Kalman Filter (KF) is integrated in order to filter the raw position estimates and make them usable by other components in the architecture.

On the side of the Long-Range UAV Detector (see Section \ref{sec:long_range}), although the position of the UAV within the image plane can be detected (a single point) , so  are unable to estimate the size of the object on the image plane . Thus, the distance of the UAV with respect to the camera frame of reference cannot be computed. In this scenario, stated distance has been fixed to a constant value, and the estimated pose is used by other modules as a waypoint that  indicates  the direction in which we need to move to get closer to the target UAV.


\subsubsection{Camera Gimbal Controller}

This component of the FES is in charge of controlling the camera gimbal in order to provide the desired behavior in every stage of the mission. The controller scheme is composed of an image-based Proportional-Derivative (PD) angular velocity controller (pitch and yaw axes). The error in the image plane is translated with geometric projections into an error in angle in camera frame of reference. Then, it is compensated by the action of this controller. This module receives two different types of modes where the FES can be in: SHORT\_RANGE\_MODE or LONG\_RANGE\_MODE. Accordingly, the behavior changes as follows:

\begin{itemize}
  \item \textit{LONG\_RANGE\_MODE:} In this mode, the camera gimbal is controlled to hold a fixed upwards pitch angle of 10$^{\circ}$. In this position, the camera gimbal is pointing towards an approximate direction of the sky in order to, first, maximize the detection probabilities of both Long-Range and Short-Range Detectors and, second, perform the grasping maneuver from a lower position with respect to the target object.
  \item \textit{SHORT\_RANGE\_MODE:} In this mode, the camera gimbal is controlled to hold the detected target (UAV or ball) in the center of the image, in order to maximize the number of frames the target object is being detected during the high-speed mission execution. The altitude of the UAV is being adjusted in parallel in order to maintain the UAV in a lower position with respect to the target ball or UAV. The control of the altitude involved a constant offset of altitude below the target UAV.
\end{itemize}

\subsubsection{Interleaved Execution Policy}\label{FES_component_flow}

\begin{figure}[h!]
    \centering
    \includegraphics[scale = 0.28]{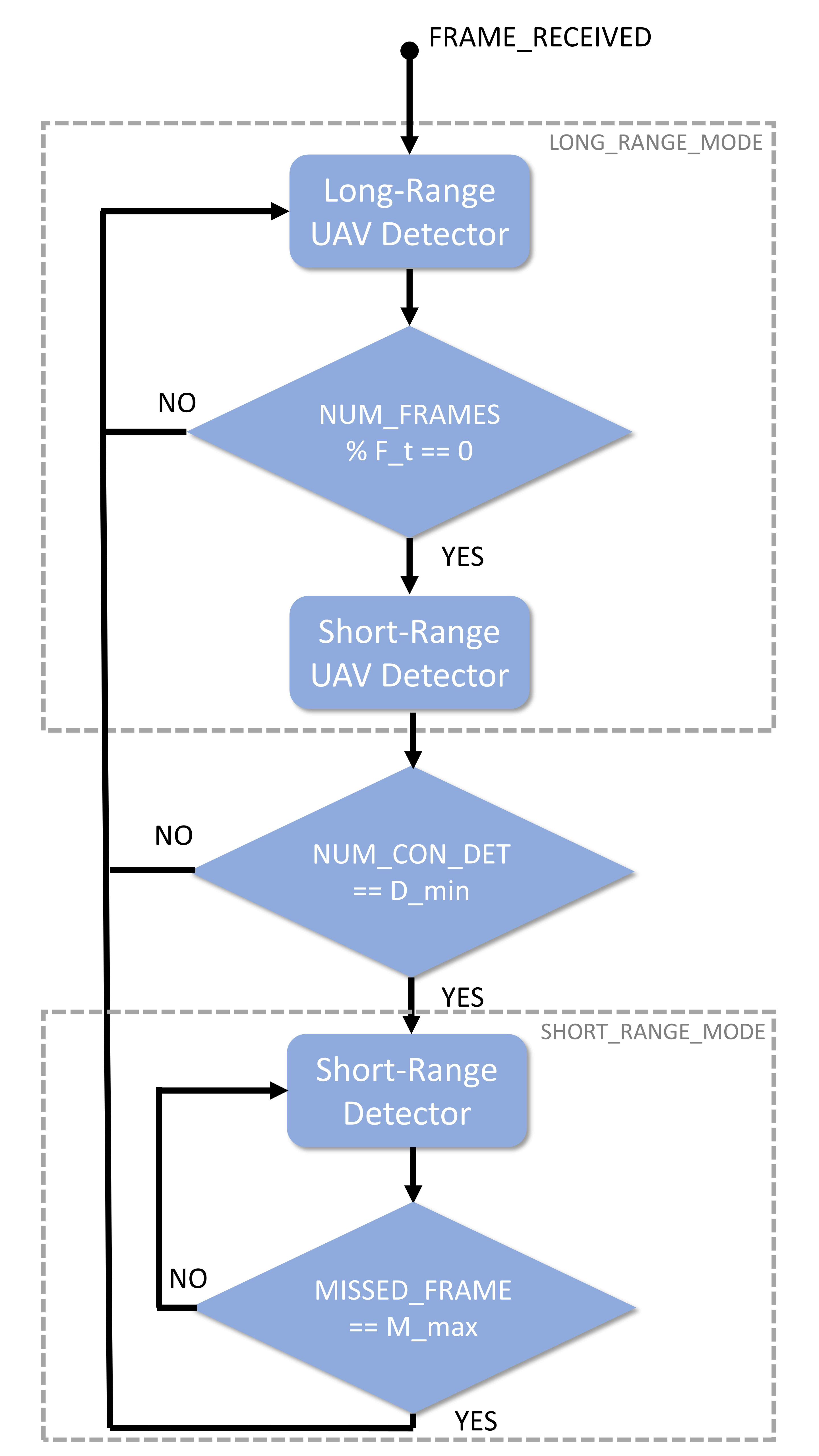}
    \caption{Feature extraction system component execution flowchart.}
    \label{fig:perception_flow} 
\end{figure}

For the sake of efficient coordination, during the execution of a mission, the FES components share a policy to decide at which time slot each detector is executed. Fig. \ref{fig:perception_flow} shows the execution flow of this policy. At the start of the mission, the Long-Range Detector is executed when an image frame is received (since the location of the target UAV is unknown and assumed at a far distance) and the amount of image frames is recorded using a counter (NUM\_FRAMES). At this moment, the FES is considered to be in LONG\_RANGE\_MODE.

After the execution of the Long-Range UAV Detector, the modulus operation is applied to NUM\_FRAMES with the divisor F\_t (number of sporadic short-range detections in between long range detections). As such, after every F\_t frames, the Short-Range Detector is executed to evaluate whether a detection (either UAV or ball detection)  is made. Once the number of consecutive frames where a short-range detection is found (NUM\_CON\_DET) equals a minimum threshold (D\_min), the Short-Range Detector is executed at every frame (FES changes to SHORT\_RANGE\_MODE) until the amount of frames without a detection (MISSED\_FRAMES) reaches a maximum value of allowed misses (M\_max). All of this constant value were experimentally defined to provide desired performance for a given experiment. It has to be highlighted that the Gimbal Controller is being continuously executed and it follows an internal policy to decide the reference angle. Also, the Position Estimator is executed every time a new detection is completed.

\subsection{State Estimation System}

Challenge 1 required our aerial robot to perform aggressive and precise maneuvers in order to intercept the object carried by the target aerial robot flying at high speeds (between 5 and 8 m/s).~Thus, fast and robust state estimation was a key element of the the autonomous system in order to perform the mission successfully. 

\begin{figure}[h!]
    \centering
    \includegraphics[width=0.75\textwidth]{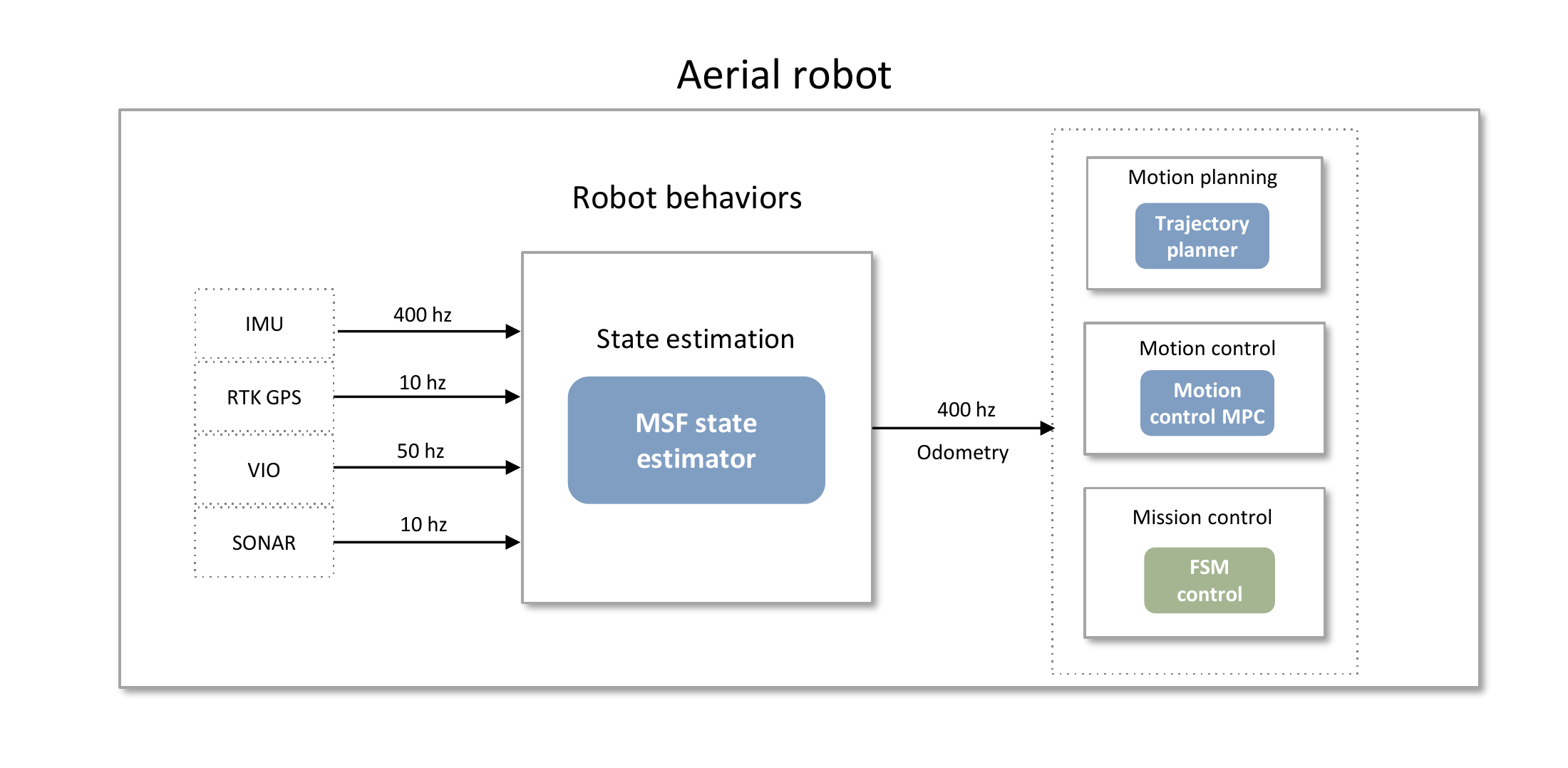}
    \caption{Block diagram for the Multi Sensor Fusion state estimation module.}
    \label{fig:MSF} 
\end{figure}

The state estimation system of our architecture, shown in Fig. \ref{fig:MSF}, consists of the robust and modular Multi-Sensor Fusion (MSF) framework in ROS \cite{ref_msf}. 
The framework is based on an Extended Kalman Filter (EKF) and has the capability to integrate an unlimited number of sensors, as well as to compensate for any delayed sensor measurements. It can also estimate the bias in the sensor measurements and allow for online self calibration of the sensor suite. 
The state prediction is performed by fusing the measurements received from the IMU sensors, which provide the measurements of linear accelerations and angular velocities in the 3 axis at a frequency of $\sim$400 Hz, and the position estimates about the x,y and z axis provided by the RTK GPS sensor at $\sim$10 Hz. For additional robustness, the framework has been modified in order to integrate the linear velocities about x, y and z axis of the aerial robot estimated by an on-board Visual Inertial Odometry (VIO) algorithm running at $\sim$50 Hz and the altitude of the aerial robot estimated by a sonar sensor. The MSF is thus able to estimate the complete high-frequency odometry information of the aerial robot at $\sim${400} Hz.

\subsection{Motion Planning System}
\label{planner}

\begin{figure}[h!]
    \centering
    \includegraphics[width=0.6\textwidth]{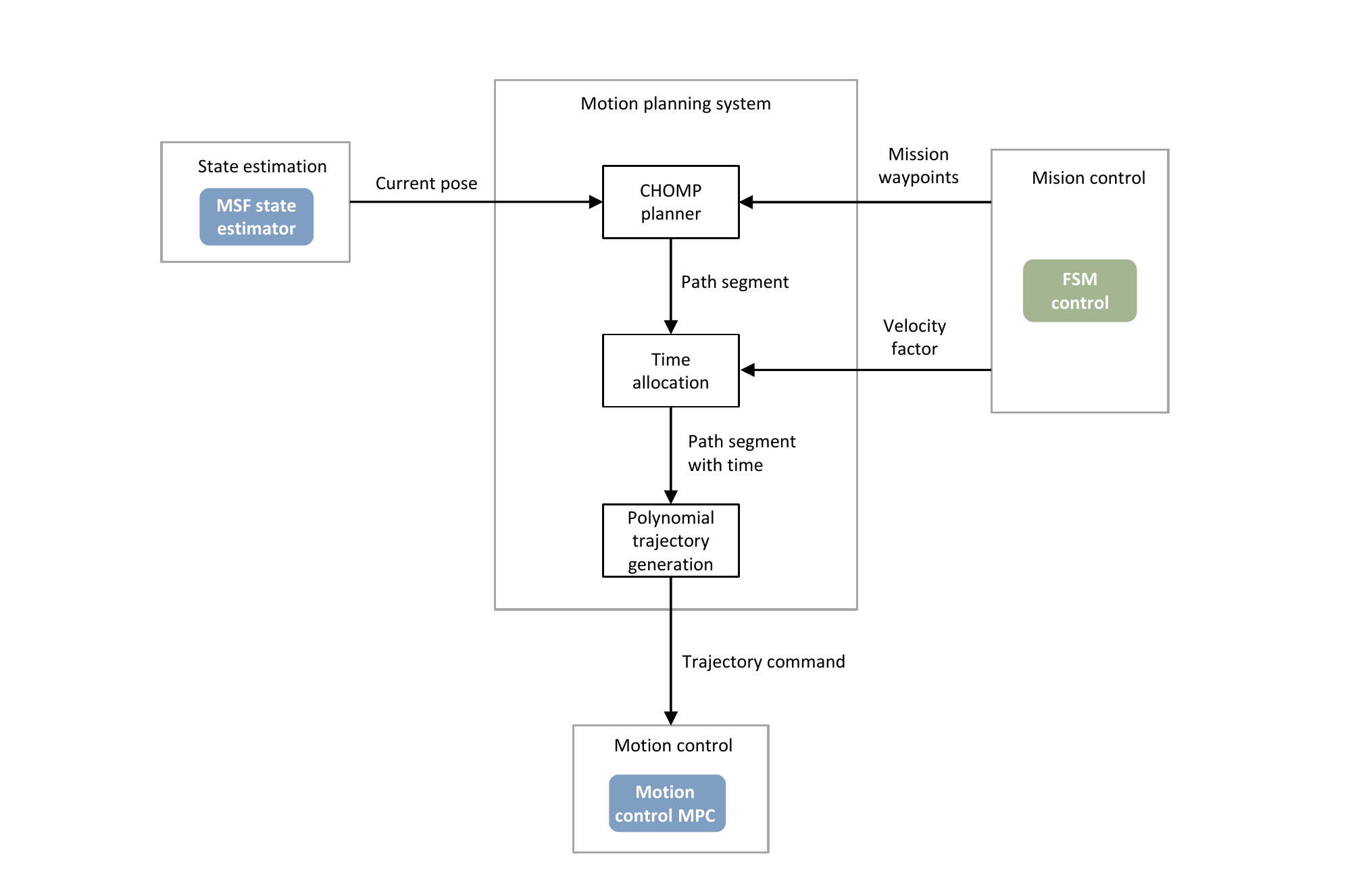}
    \caption{Schematic diagram for the motion planning system.}
    \label{fig:MPS} 
\end{figure}

The Motion Planning System is in charge of defining a trajectory along which the aerial vehicle will have to navigate in every state of the mission in order to end up performing a precise catch of the ball. A modified CHOMP planner \cite{ratliff2009chomp} is used for planning a collision-free path for catching, approaching and following the target UAV and items. The waypoints and the velocity factor for the planner are given by the mission control system, the poses of the target drone and items estimation are given by the feature extraction system, the initial point is taken as the current pose of the UAV, which is given by the MSF odometry measurements. A schematic diagram of the motion planning system can be seen in Fig. \ref{fig:MPS}.

First, the line segment between the initial and goal points is interpolated into finite points, where every point can be thought of as a state. The cost function of the CHOMP planner is composed of an obstacle objective function $f_{o}(\xi)$, a smooth objective function $f_{s}(\xi)$, the boundary objective functions $f_{b}(\xi)$ for X and Y, and the altitude objective functions $f_{a}(\xi)$ for Z. The main contribution of this equation is to generate a dynamic feasible, collision avoidance trajectory and within the defined altitude range which would be in the boundary of the flight area. In Eq. \ref{planning_eq_1}, the smooth objective functions take charge of dynamic feasibility, and the obstacle objective functions take the charge of obstacle-free. The boundary objective functions are used to penalize the trajectory if it is outside of the flight area while the altitude objective functions are used to avoid the drone flying to the ground or the ceil. The final objective function can be expressed as

\begin{equation}\label{planning_eq_1}
f(\xi) = w_{1}f_{s}(\xi) + w_{2}f_{o}(\xi) + w_{3}f_{b}(\xi) + w_{4}f_{a}(\xi)
\end{equation}
with,

\begin{equation}\label{planning_eq_2}
f_{b}(\xi) = c_{x_{max}}(\xi) + c_{x_{min}}(\xi) + c_{y_{max}}(\xi) + c_{y_{min}}(\xi)
\end{equation}

\begin{equation}\label{planning_eq_2_5}
f_{a}(\xi) = c_{z_{max}}(\xi) +c_{z_{min}}(\xi)
\end{equation}

\begin{equation}\label{planning_eq_3}
f_{s}(\xi) = \frac{1}{2}\int_{0}^{1}\|\frac{\mathrm{d}}{\mathrm{d}s}\xi(s)\|^2\mathrm{d}s
\end{equation}

\begin{equation}\label{planning_eq_4}
f_{o}(\xi) = \int_{0}^{1}c_{o}(\xi(s))\|\frac{\mathrm{d}}{\mathrm{d}s}\xi(s)\|\mathrm{d}s
\end{equation}
where $w_{1}$, $w_{2}$, $w_{3}$, and $w_{4}$ are the weights for each one of the objective functions, $\xi$ is the path and $\xi(s)$ is a function mapping the path length $s$ to the UAV configurations. In the real flight, $w_{1}$, $w_{2}$, $w_{3}$, and $w_{4}$ are set as 1, 5, 10, and 15, respectively.

Eq. \ref{planning_eq_2_5} are taken as examples to describe the boundary objective functions and the altitude objective functions. The detail about $c_{z_{max}}$ and $c_{z_{min}}$ in Eq. \ref{planning_eq_2_5} can be seen from Eq. \ref{planning_eq_2_5_1} and Eq. \ref{planning_eq_2_5_2}. $Z_{max}$ and $Z_{min}$ in Eq. \ref{planning_eq_2_5_1} and Eq. \ref{planning_eq_2_5_2} mean the maximum and minimum altitude the drone can reach while $Z(\xi(s))$ is the current altitude of the drone. $c_{x_{max}}$, $c_{x_{min}}$, $c_{y_{max}}$ and $c_{y_{min}}$ in Eq. \ref{planning_eq_2} is similar with $c_{z_{max}}$ and $c_{z_{min}}$. The only difference is that it needs to replace the Z in the formula with X or Y.

\begin{equation}\label{planning_eq_2_5_1}
c_{z_{max}}(\xi(s)) = \left\{ \begin{array}{rcl} (Z(\xi(s)) - Z_{max})^2, & & {Z(\xi(s)) \geq Z_{max}}\\ 0, & & {Z(\xi(s)) < Z_{max}} \end{array} \right.
\end{equation}

\begin{equation}\label{planning_eq_2_5_2}
    c_{z_{min}}(\xi(s)) = \left\{ \begin{array}{rcl} 0, & & {Z(\xi(s)) > Z_{min}}\\ (Z(\xi(s)) - Z_{min})^2, & & {Z(\xi(s)) \leq Z_{min}} \end{array} \right.
\end{equation}

Given that the target UAV is considered as a cylinder obstacle, in order to avoid planning paths above or below the target, the obstacle cost function parameters are expressed as,

\begin{equation}\label{planning_eq_5}
c_{o}(\xi(s)) = \left\{ \begin{array}{rcl} \frac{r(\xi(s))}{3}(1-\frac{dist(\xi(s))}{r(\xi(s))})^3, & & {dist(\xi(s)) \leq r(\xi(s))}\\ 0, & & {dist(\xi(s)) > r(\xi(s))} \end{array} \right.
\end{equation}

\begin{equation}\label{planning_eq_6}
r(\xi(s)) = \frac{R}{\cos(\arctan(\frac{dist_{z}(\xi(s), c)}{dist(\xi(s)}))}.
\end{equation}

In (\ref{planning_eq_5}) and (\ref{planning_eq_6}), the distance from current position of the UAV to center of the cylinder obstacle is denoted by $dist(\xi(s))$, $R$ is the radius of the obstacle, $c$ is the position of the center of the obstacle, and $dist_{z}(\xi(s),c)$ is the altitude distance from the current position of the UAV to the center of the obstacle. The pose and radius estimations of the target UAV and item are obtained from the feature extract system. $c_{o}(\xi(s))$ is set as 0 when the target drone pose estimation is not given. 





\begin{figure}[h!]
    \centering
    \includegraphics[width=0.45\textwidth]{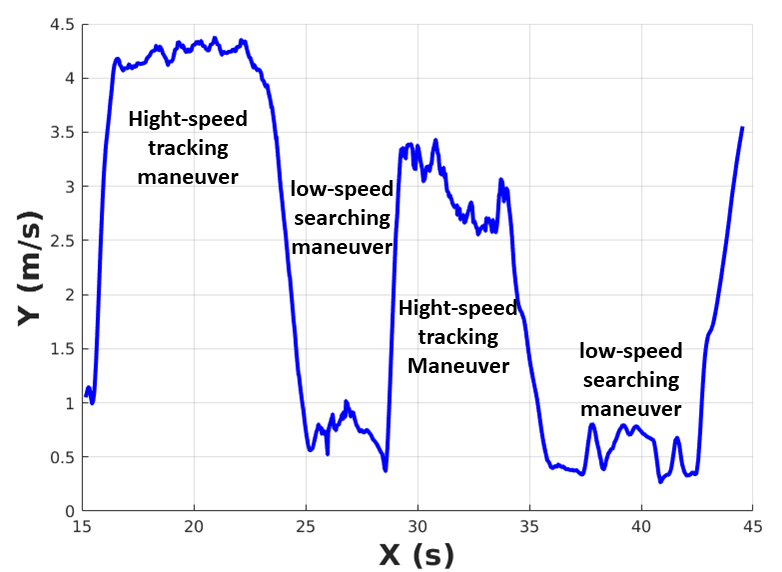}
    \caption{The velocity-changing during the high-speed tracking maneuvers and low-speed searching maneuvers. The velocity factors are set as 4 for high-speed tracking and 1 for low-speed searching, respectively.}
    \label{fig:fv} 
\end{figure}

With these functions, the cost functional gradients of every state is calculated and a gradient descent approach is used to do path optimization. Afterwards, a ROS trajectory generation package \cite{richter2016polynomial} is used to generate the desired trajectory to the target UAV or item, given the path of the CHOMP planner. In order to generate a the trajectory tracking strategy instead of path following, different velocities for certain stages of the mission were defined. With the following function the segment times were modified:

\begin{equation}
T_{seg}(\xi_{k-1}, \xi_{k}) = \frac{d \times (1 + f_{maxv} \times v_{max})}{2 \times v_{max} \times f_v \times a_{max} \times e^{\frac{-d}{2 \times v_{max}}}}
\end{equation}
where $T_{seg}(\xi_{k-1}, \xi_{k})$ is the segment time between $\xi_{k-1}$ and $\xi_{k}$, $d$ is the distance from $\xi_{k-1}$ to $\xi_{k}$, $f_{maxv}$ is the maximum velocity factor, $f_v$ is the velocity factor which is given by mission control system, $v_{max}$ and $a_{max}$ is the maximum velocity and maximum acceleration the drone can reach. Different values of velocity factors are given based on the maneuvers of fast flight with high tracking errors, slow flight for precise maneuver, and target drone searching maneuver. In the real flight, $f_{maxv}$ is set as 6.5 while $f_v$ is set as 4 for fast-tracking and 1 for flying slow to search the target and to do the precise maneuvers (searching and ball catching). The maximum velocity and maximum acceleration are set as 15 $m/s$ and 20 $m/s^2$. Fig. \ref{fig:fv} shows the velocity-changing process when switching between fast-tracking and precise maneuvers. The velocities in the graph are the sum of squares of X, Y, Z three-axis speeds.

\subsection{Motion Control System}

The Motion Control System is in charge of generating and communicating the necessary actuator commands for the UAV to perform the desired trajectories, generated by the aforementioned planning system. Given the complex requirements of Challenge 1, reliable trajectory tracking control is a key element of the overall system. To address this problem a Nonlinear Model Predictive Controller (NMPC) \cite{ref_mpc_1} has been implemented. The main advantage of MPC is the fact that it calculates the optimal control actions taking into account both the actual and future points of the trajectory within its prediction horizon and the constraints of the UAV. This is achieved by optimizing a finite time-horizon cost function, but only implementing the current control action and then optimizing again in the next iteration.

A NMPC has been implemented based on the full system dynamics of the DJI Matrice 210 v2 RTK instead of a Linear MPC which only considers a linearized model, because UAV behavior is better described by a set of nonlinear differential equations in order to capture the aerodynamic and coupling effects. As demonstrated in \cite{ref_mpc_2}, the NMPC presents a better disturbance rejection capability, step response, tracking performance and computational effort than LMPC. Thus ensuring precise trajectory tracking in realistic environments where the flight performance may be compromised by external disturbances. To enhance the predictive capability of our NMPC, an on-line EKF disturbance observer which compensated for changes in mass of the UAV and wind has been designed. 

Since the DJI Matrice 210 v2 RTK does not support directly sending angular rate commands, the MPC controller outputs are smooth pitch and roll commands for horizontal motion and altitude. Although the position and heading of the UAV can be controlled simultaneously by the MPC, these DOF have been decoupled. Thus, for simplicity, a proportional control scheme for the yaw-axis has been used. 

To configure our NMPC, all the constraints in velocities and accelerations of the UAV are indicated and the operating frequency of the control is set at 100Hz. This control signal is sent to the low level control that is done by the DJI onboard computer at 400Hz. To reduce the computational load of the NMPC, the internal prediction frequency is fixed at 10Hz following the strategy presented in \cite{ref_mpc_1}, thus, with a prediction horizon of only 20 steps, the trajectory of the next 2 seconds are optimized in the controller. As for the Q and R matrices of NMPC cost function, we set a more aggressive controller for the FOLLOW\_SHORT\_RANGE and CATCH\_BALL states, setting a more robust control for the rest of the states. This way the use of large accelerations in the most critical part of the mission is enhanced and the consumption through smaller accelerations when the UAV is far from the target is reduced.

\subsection{Mission Control System}\label{sec:executive_system}

In general, a robot mission execution can be automatically controlled with the help of a representation such as Finite State Machines (FSM), hierarchical finite state machines, behavior trees, etc. In this work, a solution based on a FSM for the architecture designed for Challenge 1 has been implemented, which is adequate according to the complexity of the problem, since it can be represented with a small number of states and events. This representation was very useful to experiment with low programming effort with different versions of the mission control. 

The implemented FSM uses a number of states in order to control different parts of the system. For example, some states activate and deactivate the Long-Range UAV Detector and Short-Range Detector. This is needed because the computation load is limited and all detectors cannot run in real time at the same time. States can also establish the required UAV speed. The UAV may need to increase the velocity in order to reach the target UAV or the ball. However, when there is not a target detected, the UAV uses low speeds to maximize the performance of the detectors. In addition, there are also states to establish navigation goals to move the UAV different positions (\textit{e.g.} to search the other drone or to reach a target).

The following list summarizes the most important states used in the FSM:

\begin{itemize}

  \item \textit{START\_STATE}: The UAV is landed. In this state, the GPS, RTK and cameras are initiated. 

  \item \textit{SEARCH}: The UAV does a predefined trajectory which was designed to maximize the probability to detect the target UAV. The trajectory is in the form of a semi ellipse, very close to the long axis border and in the middle of this axis. This is because it is known that the lemniscate will cover 80\% of the arena's, therefore the centre of it should be situated in the center part of the arena and this trajectory put this area in the FOV of the camera.Also, the trajectory has a low altitude, so the camera is pointing to the sky. This maximizes the accuracy of the detectors because the background is uniform and there are fewer objects that can move. Finally, during this state, the manoeuvres are at very low speed and acceleration order to get sharp images. In this state both UAV detectors are active
  
  \item \textit{FOLLOW\_LONG\_RANGE}: Every time there is a new long-range detection the UAV moves at a high speed to a point  5 meters towards the detection in the XY plane, while the altitude of the UAV is kept at a constant and low value in order to maximize the performance of the detectors.
  
  \item \textit{FOLLOW\_SHORT\_RANGE}: The UAV moves at a high speed towards the detection, maintaining 4 meters from the target. In this state, the Long-Range UAV Detector is disabled so the Short-Range Detector take all the computation power. 
  
  \item \textit{CATCH\_BALL}: The  UAV moves towards the ball detection trying to align the ball position with the gripper. Once the ball goes out of the camera field of view, it continues following the planned trajectory. In this state, the trajectory planner generates the required movements and the Long-Range UAV Detector is disabled so that the Short-Range Detector takes all the computation power. 
  
  \item \textit{LAND}: The UAV lands to a given position and the mission ends.
  
\end{itemize}

\begin{figure}[h!]
    \centering
    \includegraphics[width=0.70\textwidth]{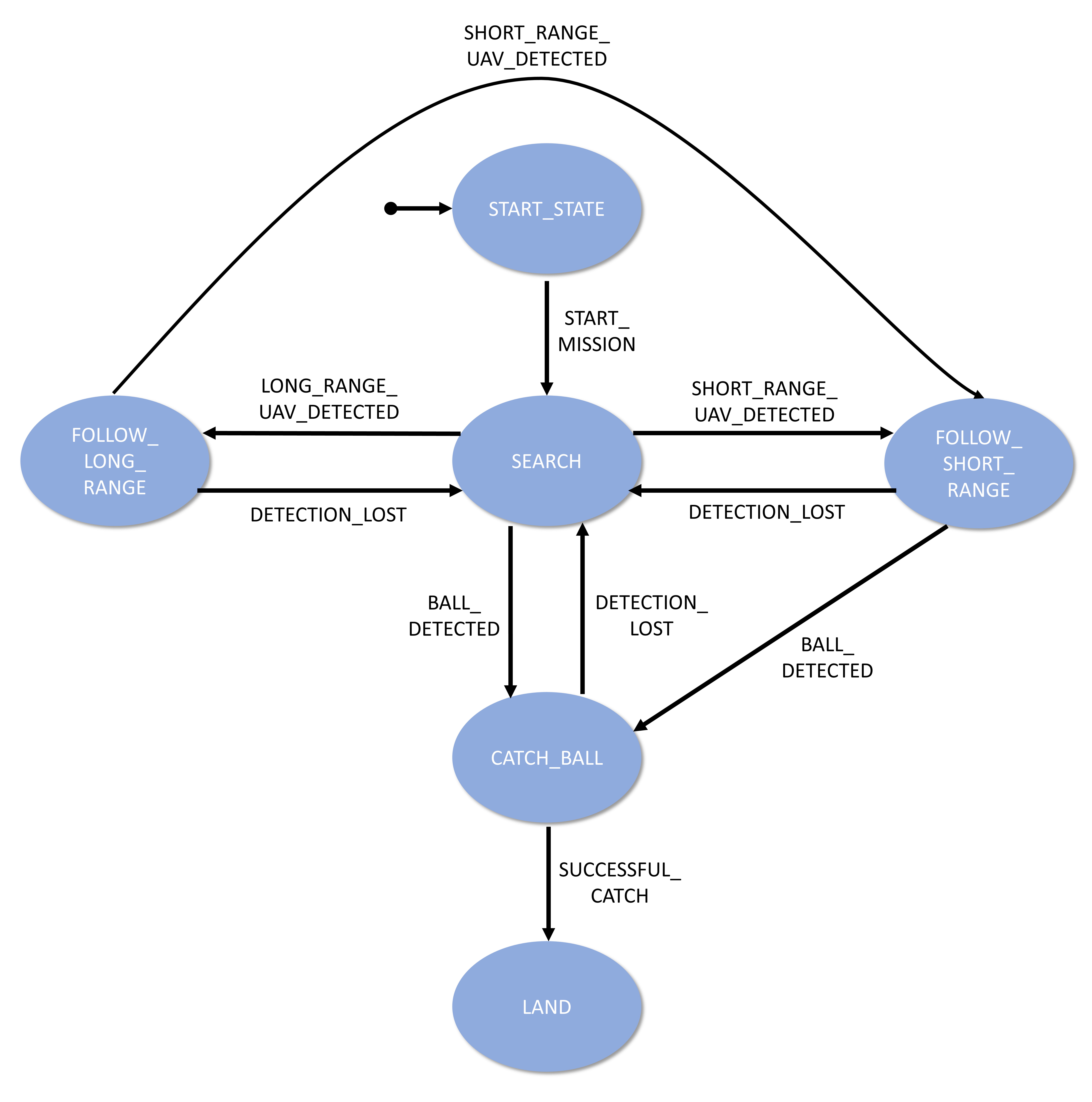}
    \caption{View of the states and transitions of the finite state machine.}
    \label{fig:finite_state_machine} 
\end{figure}

Table \ref{tab:events_fsm} shows a list of example events that were used in the FSM as transitions between states. Some of these events are related to the perception performance considering the number of consecutive image frames in which an object is detected. For example, a short range detection is generated when the target UAV is recognized in 2 frames from the last 4 consecutive frames. Figure \ref{fig:finite_state_machine} illustrates part of the transitions in the FSM.

\begin{table}[h]
  \centering
  \caption{\label{tab:events_fsm} Example events used in the finite state machine.}
  \label{tab:long_range_results}
  \begin{tabular}{  p{6cm} | p{8cm} } 
    \toprule
    \textbf{Event} & \textbf{Description} \\
    \hline
    START\_MISSION & The human operator gives the order to the UAV to start the mission \\
\hline
LONG\_RANGE\_UAV\_DETECTED & There are  3 detections on 5 consecutive frames from the Long-Range Detector. \\
\hline
SHORT\_RANGE\_UAV\_DETECTED  & There are 2 detections on 4 consecutive frames from the Short-Range Detector. \\
\hline
BALL\_DETECTED & The ball is detected. \\
\hline
DETECTION\_LOST & No new detections in 5 consecutive frames \\
\hline
SUCCESSFUL\_CATCH & Ball is detected by the laser sensor inside gripper. \\\bottomrule
  \end{tabular}
\end{table}


\section{Experiments and Results}
\label{sec:experiments_and_res}

In this section, several experimental tests have been performed to test and validate all of the proposed solutions which have been integrated in the system pipelines. In this section, the major results achieved for challenges 1 are presented.

\subsection{Experimental Setup}

Since the challenge was to be performed in an outdoor environment, an open airfield provided a workspace large enough (larger than 100 m $\times$ 40 m $\times$ 20 m) to test several scenarios that could be encountered during the real competition trials. The initial tests performed were focused on gradually validating the individual components of the system architecture, presented in Section \ref{sec:system_description}. Nevertheless, the experiments and tests provided in this section correspond to the report of a global mission execution.

In order to perform the experiments, a DJI Matrice 100 is commanded to perform a lemniscate reference trajectory (of approximate size 80\% the total allowed area for the challenge) at different velocities and at 15 m reference altitude, replicating the challenge environment. The DJI Matrice 100 carried a yellow balloon, tethered to a carbon fiber stick of approximately 1.5 m in length.

All the computation for the DJI Matrice 210 v2 RTK is performed on-board, with the hardware setup described in Section \ref{sec:hw_setup} and the mission is fully autonomous. A safety pilot has been constantly aware of the state of the mission execution in order to avoid risks in case of failure.

\subsection{Training and Test of the Feature Extractors}
\label{sec:fes_results}

There are two main feature extractors, which are part of the FES, and are in charge of aiding both the Long-Range UAV Detector and the Short-Range Detector.

\subsubsection{Training Methodology}

The training methodology is different in every detector case, despite being based on CNN architectures. The main aspects are as follows:

\begin{itemize}
    \item \textit{Long-Range UAV Detector:} This component is built on several processing steps. One of them involves the generation of a CNN architecture for UAV/Background classification. This model has been trained with a binary cross-entropy loss and Adam optimizer with learning rate of 0.001. Every batch during the training phase has been composed of a 50\%-50\% balanced UAV-background patch samples, in order to stabilize training due to the high amount of background samples in the original dataset. The batch size has been 1024 and dropout has been included in every fully connected layer. The dataset used for training has been the Purdue UAV Dataset\footnote{\url{https://engineering.purdue.edu/~bouman/UAV_Dataset/}}.
    \item \textit{Short-Range Detector:} This component consists on a single neural network that performs the detection on a single inference pass. The training procedure consists on three steps: 
    \begin{enumerate}
        \item \textit{Imagenet pretraining:} In order to get rich features and lower the training time, we use a pretrained  darknet53  \cite{ref_yolo} on Imagenet. We copy the weights from darknet53 to the matching layers of tiny YOLOv3. 
        \item \textit{General detector training:} The network is trained  using our baseline dataset. The aim of this dataset is to obtain a general detector that can detect the objects of interest( UAVs, balls and balloons) in many different situations and environments. The dataset contains almost 4000 images of different models of UAVs, and 1000 images of balloons and balls gathered online that we labeled manually. For training the network we use the Darknet \footnote{\url{https://github.com/alexeyab/darknet}} framework. We trained the network during 10000 steps using Adam optimizer , with a learning rate of 0.001, lowering it to 10\% and 1\% at 8000 and 9000 steps respectively. We used a 20\% of the training dataset for validation purposes and early stopping. The data augmentation parameters are shown in Table \ref{tab:data_augmentation_yolo}.
        \item \textit{Domain specific fine tuning:} Once the detector is performing properly, we perform a fine tuning step (lowering the initial learning rate to 0.0001) using a second dataset from the final environment. We recorded this dataset using the platform camera during the real flights using the currents objects to detect (DJI f450 and DJI m100 UAVs, green and white balloons and a yellow ball) in order to lower the domain gap between our general dataset and the real environment. Also, in order to improve the IOU of the detections, we increase the IOU (Intersection Over Union) threshold from the YOLO loss function  from 0.7 to 0.9, resulting in more accurate bounding boxes and  improving the pose estimation in subsequent steps at a marginal increase of missed detections. The data augmentation parameters are shown in Table \ref{tab:data_augmentation_yolo}.
    \end{enumerate}
\end{itemize}
\begin{table}[h]
  \centering
  \caption{Data augmentation parameters using for training the Short-Range Detector using the Darknet framework.}
  \label{tab:data_augmentation_yolo}
  \begin{tabular}{  c | c | c} 
    Parameter & General training & Fine tuning \\
    \toprule
    Saturation  & 2.0 & 1.5\\\hline
    Exposure  & 2.0 & 1.5\\\hline
    Hue  & 0.5 & 0.1\\\bottomrule
  \end{tabular}
\end{table}
\subsubsection{Feature extractors results}
\begin{figure}[h!]
	\centering
	\subfigure[]{\includegraphics[width=0.49\textwidth]{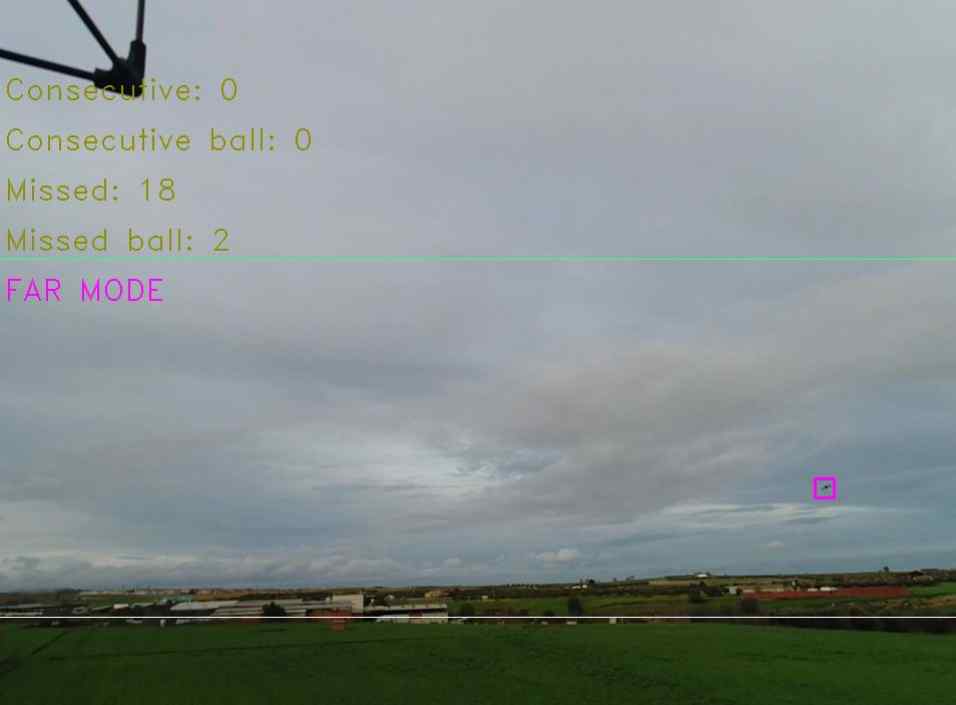}\label{far_detector_1}}
	\subfigure[]{\includegraphics[width=0.49\textwidth]{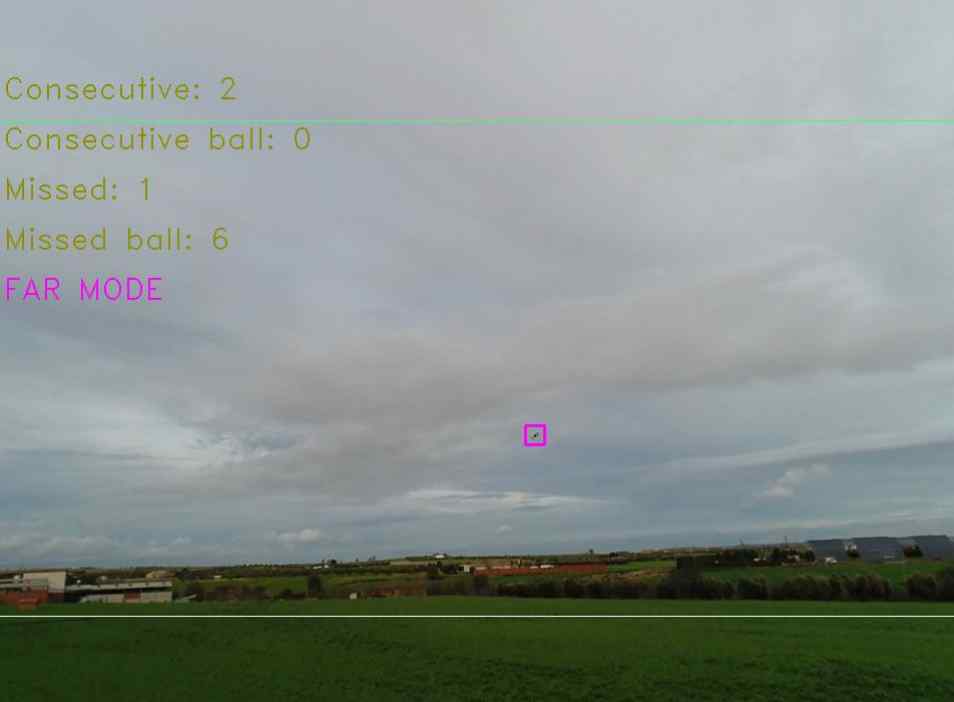}\label{far_detector_2}}
	\caption{Target UAV detection results from the Long-Range UAV Detector module. The UAV detections are depicted in magenta.}
	\label{fig:far_detector_exp}
\end{figure}

The FES modules are considered key elements in this work and they are required to successfully accomplish the tasks proposed in the challenge. As such, extensive testing was carried out to evaluate the different components of the FES. In Fig. \ref{fig:far_detector_exp}, successful detections of the target UAV moving in far distance for the camera are depicted.

The aerial vehicle was initially flying to a distance of $\sim$50 m from the target UAV. As can be seen in Fig. \ref{far_detector_1} and Fig. \ref{far_detector_2}, at this distance the Far detector was able to successfully detect the target UAV in the image frame and an approximate pose was obtained, where the magenta Region-Of-Interest (ROI) indicates the detection of the target UAV in the image frame. 

Furthermore, in order to formally evaluate the performance of the CNN classifier of the Long-Range UAV Detector component, average Frequency (Hz), Precision, Recall and F-score are provided in Table \ref{tab:long_range_results}. The provided frequency value corresponds to the complete Long-Range UAV Detection component, that is, all the processing steps described in Section \ref{sec:long_range}.

\begin{table}[h]
  \centering
  \caption{Metrics for the CNN classifier of the Long-Range UAV Detector. Precision, Recall, F-score and average Frequency (Hz) results are provided.}
  \label{tab:long_range_results}
  \begin{tabular}{  c | c } 
    \toprule
    Precision & 0.477 \\\hline
    Recall & 0.497 \\\hline
    F1-score & 0.487 \\\hline
    Frequency (Hz) & 15.2 \\\bottomrule
  \end{tabular}
\end{table}

\begin{table}[h]
  \centering
  \caption{Metrics of the final Short-Range Detector for a 0.25 confidence threshold and 0.5 IOU threshold. }
  \label{tab:metrics_YOLO}
  \begin{tabular}{  c | c } 
    \toprule
    AP (uav) & 0.893 \\\hline
    AP (ball) & 0.922 \\\hline
    AP (ballon) & 0.924 \\\hline
    mAP  & 0.912 \\\hline
    Average IOU  & 0.573 \\\hline
    Precision  & 0.9 \\\hline
    Recall & 0.81 \\\hline
    F1-score  & 0.86 \\\hline
    Frequency (Hz) & 22.54 \\\bottomrule
      \end{tabular}
\end{table}

\begin{figure}[h!]
	\centering
	\subfigure[]{\includegraphics[width=0.49\textwidth]{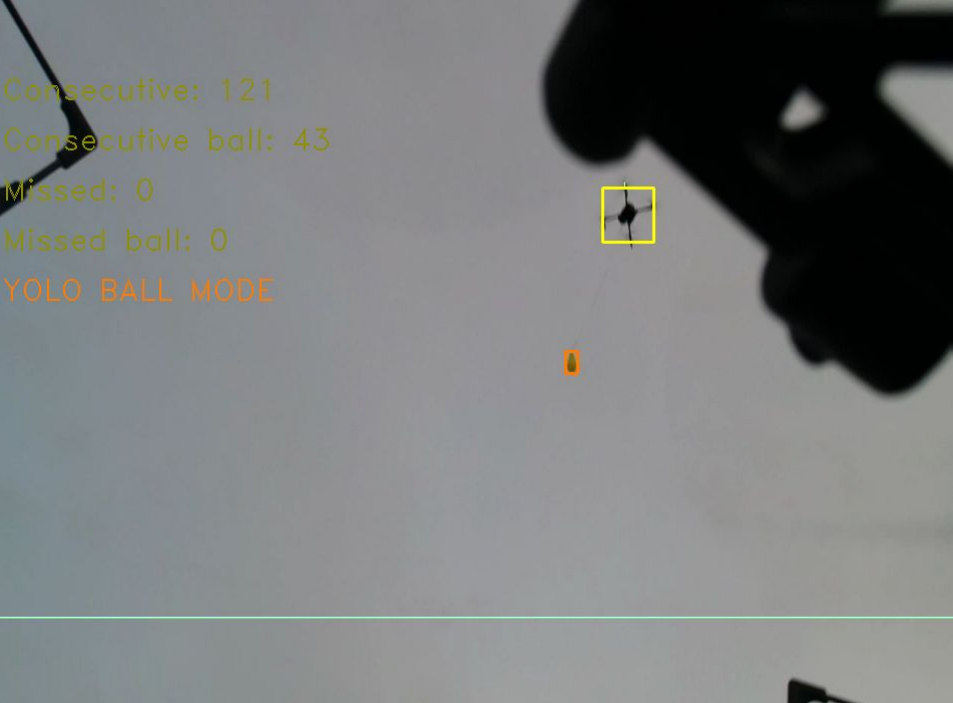}\label{close_detector_2}}
	\subfigure[]{\includegraphics[width=0.49\textwidth]{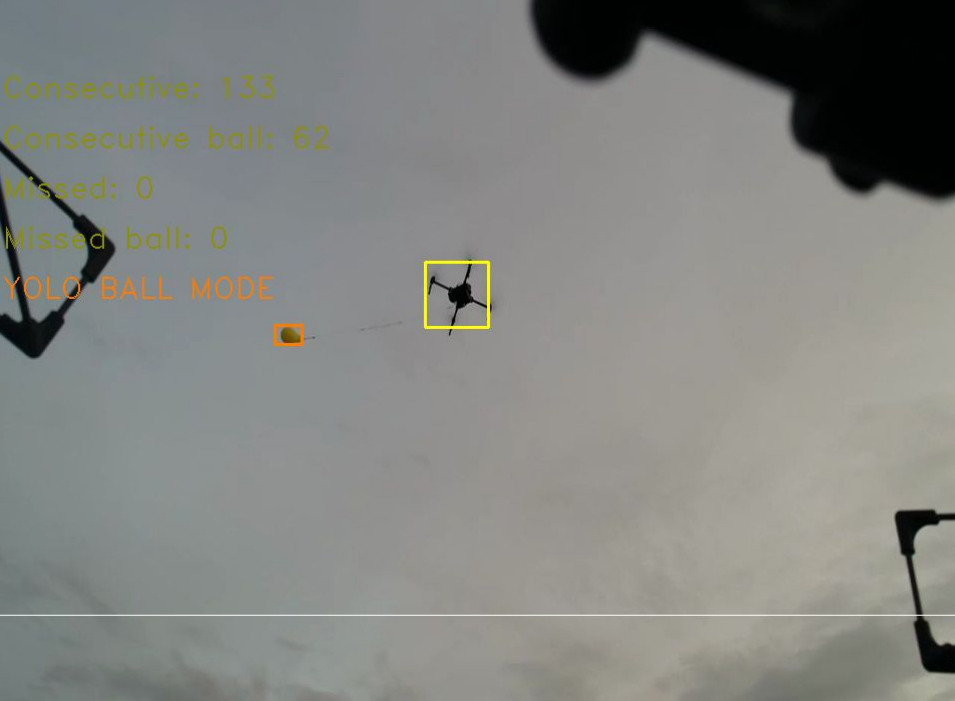}\label{close_detector_3}}
	\caption{Target UAV and Item detection results from the Close and Item detector module.}
	\label{fig:close_detector_exp}
\end{figure}

In Fig. \ref{fig:close_detector_exp}, successful detection of both the target UAV and the object (ball), in a closer distance ($\sim$10 m) are shown. The Short-Range Detector was able to detect and identify both the target UAV and the object (ball) tethered to the vehicle. In these tests, the aerial vehicle was manually flown towards and away from the target UAV, where the FES components would activate depending on the detector execution flowchart presented in Section \ref{FES_component_flow}. 

\begin{figure}[h!]
	\centering
	\subfigure[]{\includegraphics[height = 5.0 cm,width=0.49\textwidth]{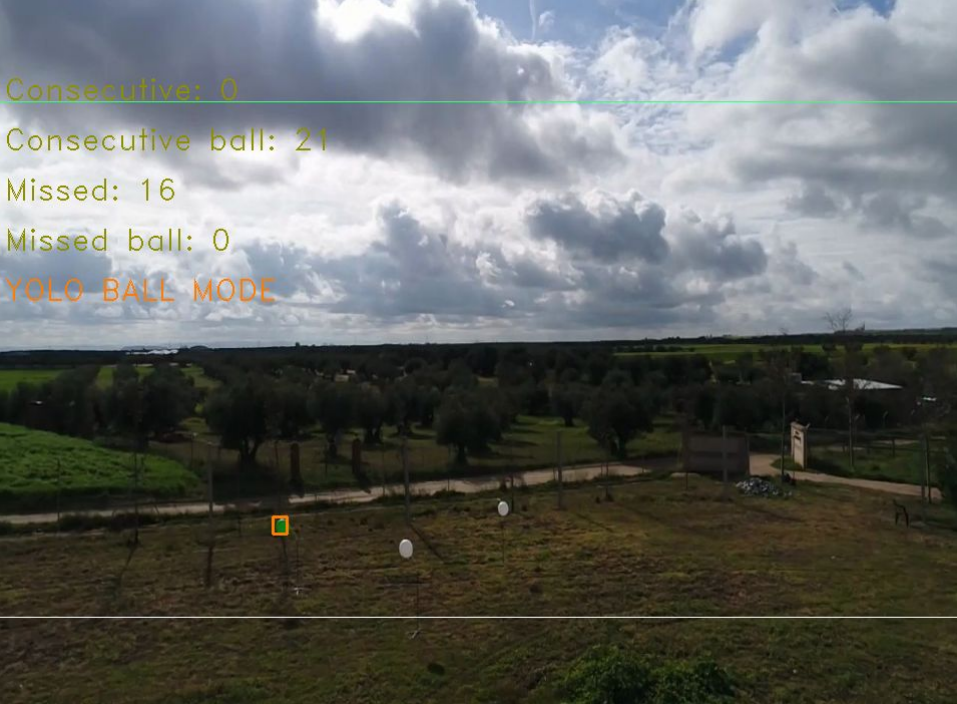}\label{balloon_detector_1}}
	\subfigure[]{\includegraphics[height = 5.0 cm,width=0.49\textwidth]{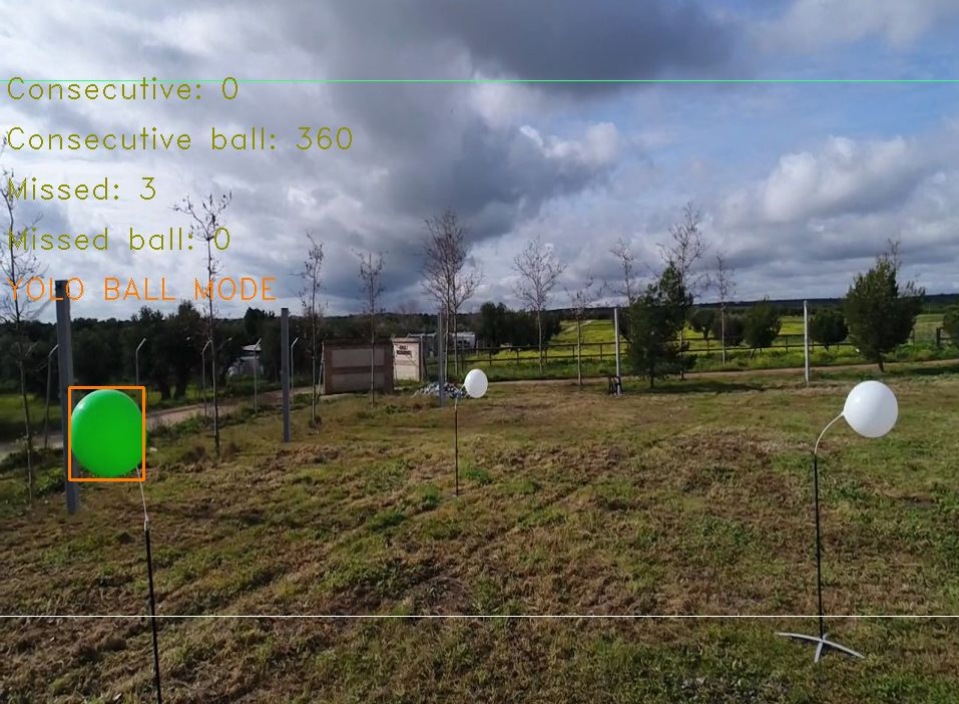}\label{balloon_detector_2}}
	\caption{Balloon item detection using the Close and Item detector module.}
	\label{fig:balloon_detector_exp}
\end{figure}

As additional results, a complementary task for the challenge was required, and involved the autonomous bursting of ground-tethered balloons. As shown in Fig. \ref{fig:balloon_detector_exp}, the balloons were properly detected by the Short-Range Detector.

In Table \ref{tab:metrics_YOLO} we report common object detection metrics for the Short-Range Detector. The AP (Average Precision)   is calculated using a 0.5 IOU threshold, while the precision and recall are calculated using a 0.25 confidence threshold.

\subsection{Target Search, Follow and Approach Results}
\label{sec:search_follow}

\begin{figure}[h!]
	\centering
	\includegraphics[width=0.48\textwidth]{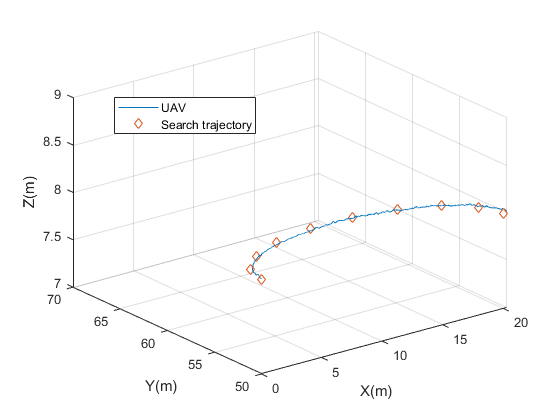}\label{close_detector_1}

	\caption{Trajectory tracking results for the SEARCH state. This is a predefined trajectory where the follower UAV tries to detect in LONG\_RANGE or SHORT\_RANGE the target UAV.}
	\label{fig:trajectory_search}
\end{figure}

 The technical specifications for Challenge 1 stated that the target UAV was flying in a lemniscate curve at a maximum velocity of 8 m/s. Thus, in order to perform the SEARCH maneuver, the Motion Planning was used to command a trajectory to the Motion Control from a specified series of waypoints following a semi-ellipsoidal shape in the x-y plane (see Fig. \ref{fig:trajectory_search}).   

The results of the trajectory tracking test are presented in Fig. \ref{fig:trajectory_search}. As previously mentioned, the MSF module from the FES is in charge of fusing the measurements from the RTK, VIO and IMU sensors to output an estimation of the pose of the aerial vehicle. Using the fused measurements from the MSF module, the MPC module of the Motion Control was able to perform the desired flight trajectories even in non-ideal conditions compensating for wind gusts of up to 15 m/s. 

\begin{figure}[h!]
	\centering
	\subfigure[]{\includegraphics[width=0.48\textwidth]{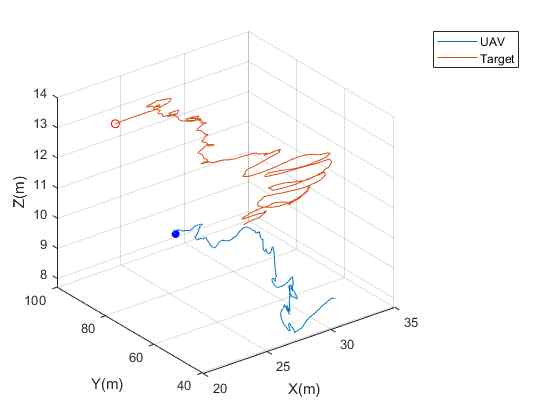}\label{close_detector_1}}
	\subfigure[]{\includegraphics[width=0.48\textwidth]{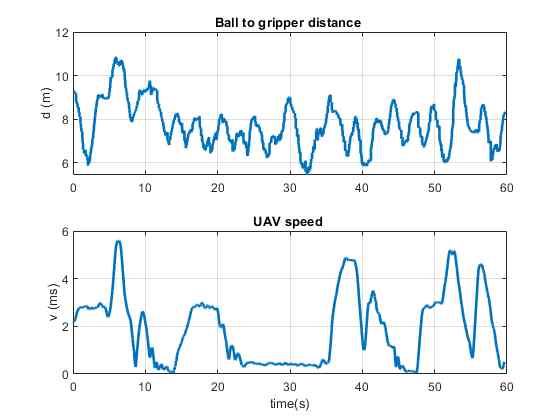}\label{close_detector_2}}
	\caption{Trajectory tracking results for the FOLLOW SHORT RANGE state. (a) trajectories of the target UAV and our proposed UAV. (b) distance between the target UAV and our proposed UAV and velocity of our proposed UAV. }
	\label{fig:trajectory_follow}
\end{figure}

In the following Fig. \ref{fig:trajectory_follow}, the FOLLOW SHORT RANGE stage is depicted. The figure shows the attempted evasion of the target UAV (see  Fig. \ref{close_detector_1}) and the maneuver of the proposed UAV to maintain a distance of about 8 meters at all moments by applying an average speed of 2 m/s (see Fig. \ref{close_detector_2}). It should be noted that at all times a security distance is maintained with respect to the target UAV when the distance is less than 6 meters, reducing the speed significantly (instants of time from 25 to 35 s).

\subsection{Grasping Maneuver Results}
\label{sec:grasping_mane}

The most challenging task of Challenge 1 is to grasp the item which is tethered to the target UAV in motion. Therefore, a test scenario was implemented to validate the performance of the overall proposed strategy in a fully autonomous real-world grasping maneuver mission. 

In this scenario, a DJI Matrice M100 UAV was used as a target UAV which has a yellow ball (a balloon for safety) tethered to the end of a 2 m long carbon fiber rod. As in the actual mission, the aerial platform is required to search, detect, follow and interact with the target UAV in order to retrieve the desired item. 

\begin{figure}[h!]
    \centering
    \subfigure[]{\includegraphics[height = 3.2 cm,width=0.49\textwidth]{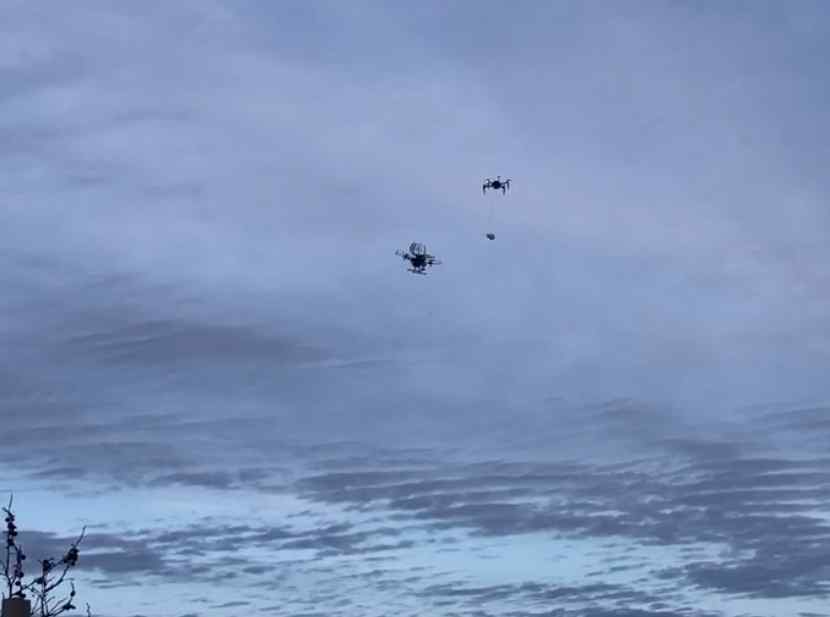}\label{fig:ball_catching_1}}
    \subfigure[]{\includegraphics[height = 3.2 cm,width=0.49\textwidth]{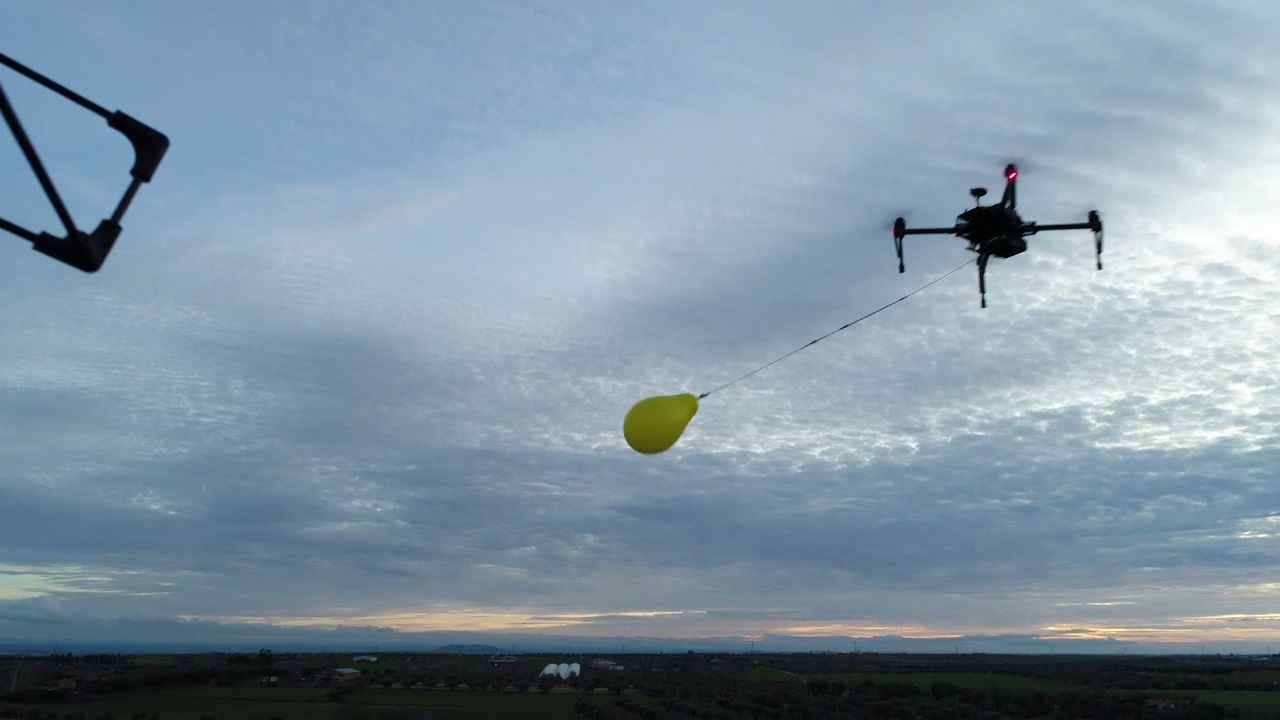}\label{fig:ball_catching_2}}
    \subfigure[]{\includegraphics[height = 3.2 cm,width=0.49\textwidth]{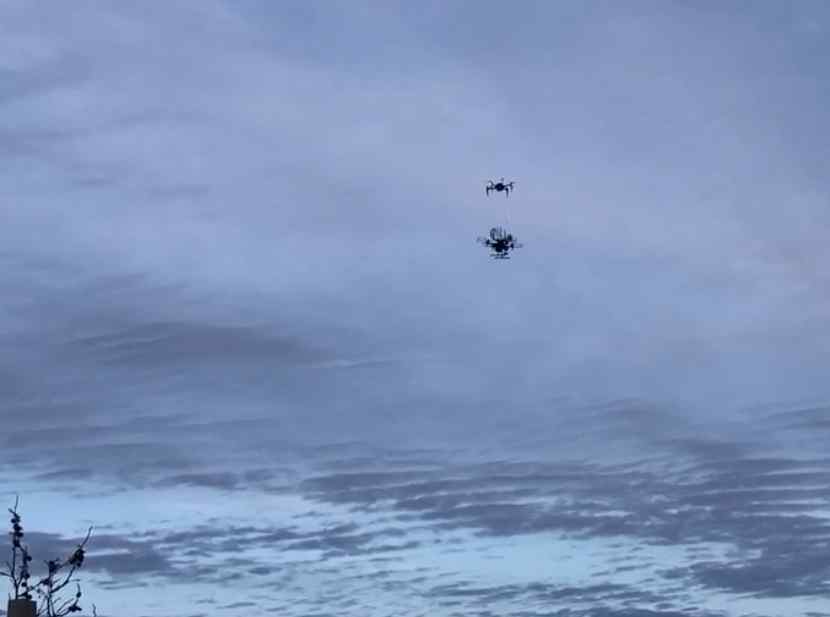}\label{fig:ball_catching_3}}
    \subfigure[]{\includegraphics[height = 3.2 cm,width=0.49\textwidth]{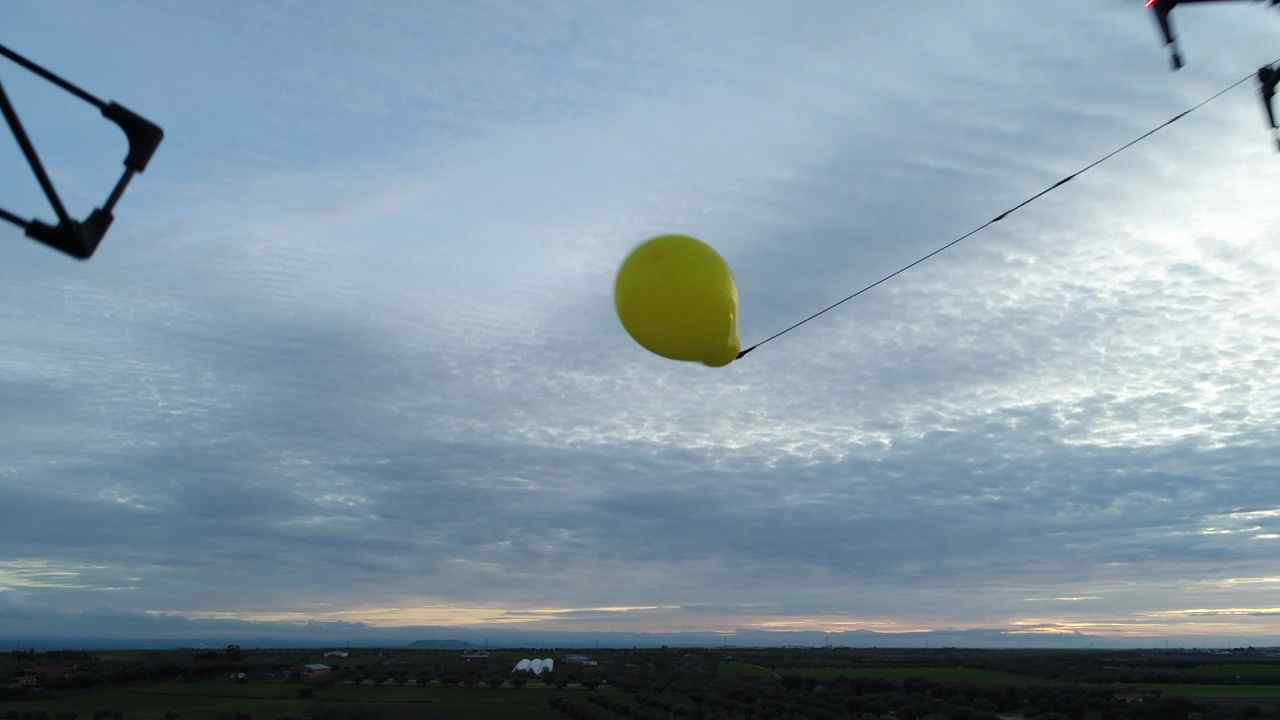}\label{fig:ball_catching_4}}
    \subfigure[]{\includegraphics[height = 3.2 cm,width=0.49\textwidth]{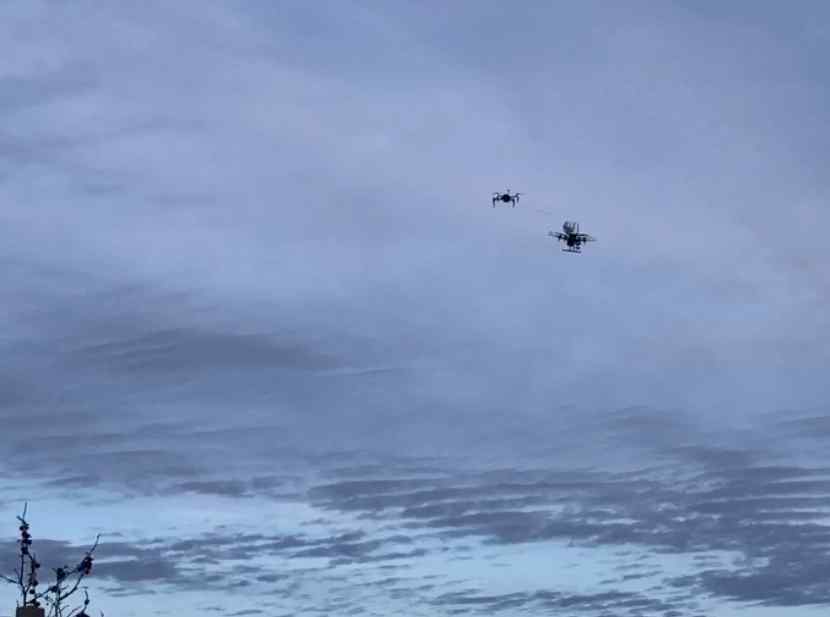}\label{fig:ball_catching_5}}
    \subfigure[]{\includegraphics[height = 3.2 cm,width=0.49\textwidth]{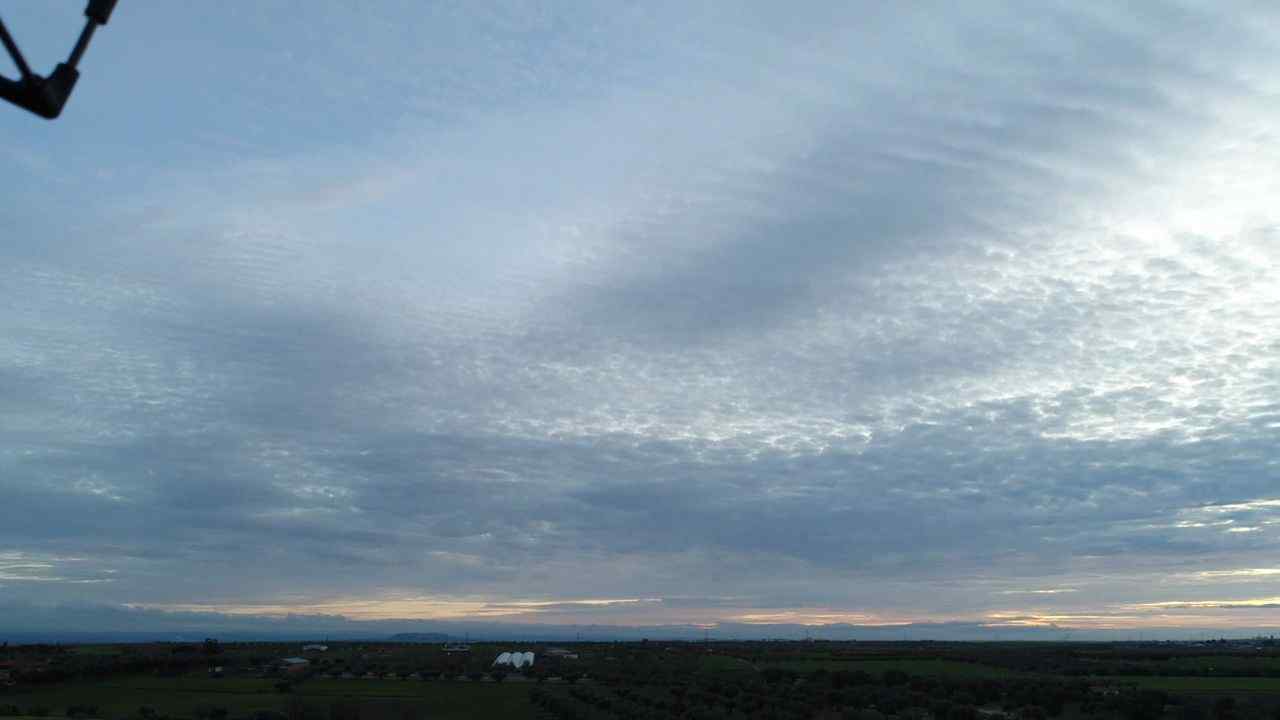}\label{fig:ball_catching_6}}

    \caption{Results from the grasping maneuver test. (a)(c)(e) Sequence view of the aerial vehicle grasping maneuver. (b)(d)(f) Images from the on-board camera used for detection of the target UAV and item.}
    \label{fig:ball_catching} 
\end{figure}
\begin{figure}[h!]
 \centering
    \includegraphics[width=0.95\textwidth]{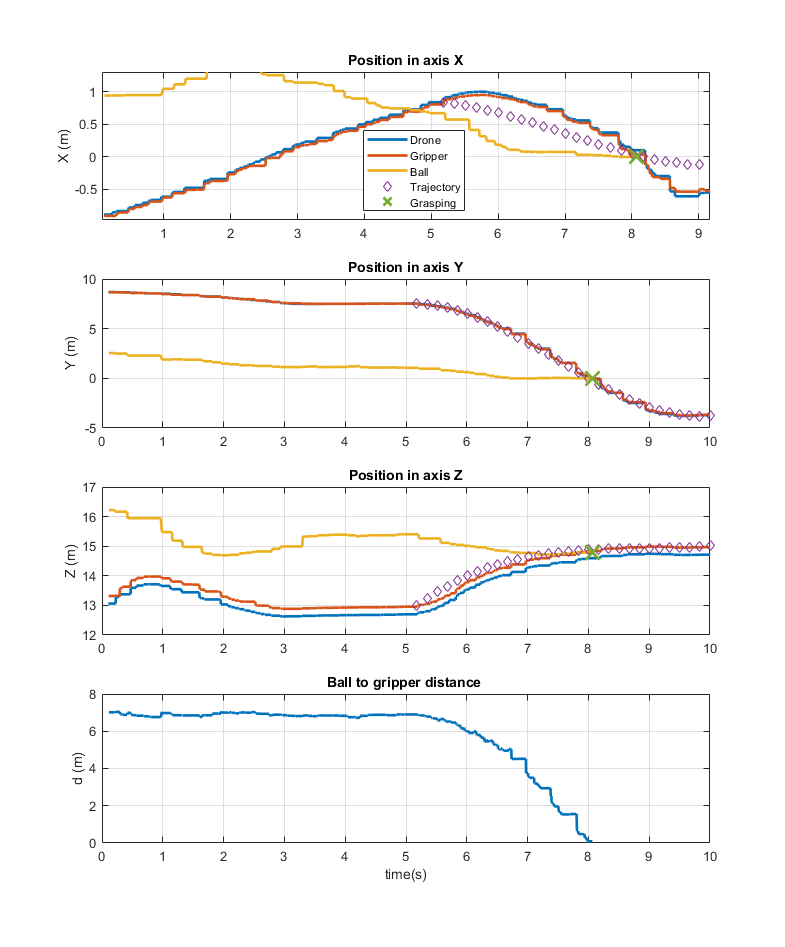}
     \caption{Trajectories of our UAV, gripper, and ball in x, y, and z axes for the grasping maneuver test during a real flight. At the beginning of the test, our UAV is following the ball keeping a distance of approximately 7 meters. At 5.1 seconds our UAV starts the grasping manoeuver, approximately 3 seconds later the distance between the gripper and the ball is reduced to 0, meaning that the grasping manoeuver was successful.}
    \label{fig:ball_catching_7}
    \end{figure}

First, the target UAV was flown to a distance of approximately 50 m from the aerial platform in flight. Afterwards, the FSM implemented in the Mission Control (presented in Section \ref{sec:executive_system}) was activated to perform the fully autonomous mission. The first states that were executed performed a grid search of the target UAV in the open airspace. Once the Long-Range UAV Detector achieved the minimum number of detections necessary, the aerial vehicle proceeded to move towards the target UAV in order to acquire a lock with the Short-Range Detector.

Once the Short-Range Detector obtained a desired number of consecutive detections, the estimated pose of the tethered object was continuously commanded to the Motion Planning system as to generate the desired trajectories needed to intersect the object with the gripping mechanism. The results of the grasping maneuver test performed with the proposed system strategy are presented in Fig. \ref{fig:ball_catching}. Fig. \ref{fig:ball_catching_1} - \ref{fig:ball_catching_6} show a sequence of images of the aerial vehicle performing the item grasping maneuver during the test flight. Fig. \ref{fig:ball_catching_7} shows the trajectory of the aerial vehicle and pose of the ball object. As can be seen, the overall autonomous system strategy proposed and implemented for the grasping task of Challenge 1 is able to retrieve the item from the target UAV successfully. Fig. \ref{fig:ball_catching} shows how the FOLLOW state is maintained until the moment of time 5.2 s. In that instant, CATCH BALL state is activated and the trajectory is planned for the gripper to catch the ball at the time instant 8 s.


\section{Discussion}
\label{sec:discussion}
An extensive validation of the proposed hardware and software UAV platform has been thoroughly carried out. All of the components have been detailed and a global mission has been performed and validated, with a separate explanation of the results. From a global perspective, the proposed UAV platform has shown outstanding results, being able to grasp a tethered ball from a moving UAV at approximately 6 m/s and resulting in the 3$^{rd}$ position for the Grand Challenge in the 2020 MBZIRC.

As explained in previous sections, the complete approach has been mainly based on the RGB camera gimbal to support the global mission. In this regard, the FES components have provided successful results. The test metrics provided in Table \ref{tab:long_range_results} and \ref{tab:metrics_YOLO} indicate a proper performance of the components. Also, the results obtained in Section \ref{sec:fes_results}, \ref{sec:search_follow} and \ref{sec:grasping_mane} demonstrate that the perception components developed for the FES are capable of detecting the target UAV and objects with varying lighting conditions, in addition to validating the FES component execution and pose estimation modules. It has to be noted that every image detection lead to a consequent position estimation, which is directly forwarded to the rest of the modules in order to complete the execution of the mission. Although the FES components provided adequate performance, the RGB camera gimbal resolution limited the behavior. For instance, an optical-zoom RGB camera gimbal can provide a wider versatility. 

Regarding the search, follow and approach results, all of the components involved showed an adequate behavior, being run on-board with tight computational resources. As shown in Fig. \ref{fig:trajectory_search} and \ref{fig:trajectory_follow}, the target UAV has been located during the SEARCH stage and latterly followed during the FOLLOW and APPROACH states. Although a safety distance has been maintaned during its execution, the target UAV has been continuously followed during the whole trajectory at high speeds. The target UAV has been followed with peak velocities of approximately 6 m/s, with on-board and vision-based components. In the different trials, the target UAV can be missed and the FOLLOW and APPROACH can be switched to the SEARCH state. This is mostly due to errors in detection or position estimation, and/or to abrupt changes in the target trajectory which hardens the motion control.

Finally, considering the grasping maneuver, the tethered ball was successfully caught. Fig. \ref{fig:ball_catching} shows the proper approximation of the proposed UAV, decreasing the distance in every axis. There is a period of time where the RGB camera gimbal is not able to perceive the ball, due to the ball and camera relative positions (close to the end of the maneuver). At this stage, the position is being actively estimated in order to enable the complete catching maneuver. As shown, the distance to the ball is decreased to approximately zero, and the ball is properly grasped at high speeds and in a continuous trajectory trend. The grasping maneuver can fail in a trial, mostly due to detections or position estimation errors and/or sudden movements of the target ball, which complicates the maneuver.

\section{Conclusions}
\label{sec:conclusions}

In this work, the hardware and software components which were implemented as the proposed strategies to solve the Challenge 1 of the 2020 Mohamed Bin Zayed International Robotics Challenge (MBZIRC) have been detailed. The team has achieved 3$^{rd}$ position for the Grand Challenge. The proposed challenges presented complex environments and scenarios, which were required to be completed using fully autonomous robotic systems. As such, a diverse knowledge of various fields has been necessary to implement efficient and robust solutions.

The Challenge 1 has been solved by means of a complete UAV platform with custom software and gripper actuator designs. All the software components has been launched on-board with tight computational budgets and primarily with vision-based techniques. The experiments have been carried out in an outdoor space, replicating the competition environment and constraints. The whole challenge mission has been successfully executed and detailed, providing adequate results for every component of the hardware and software architecture, and being able to grasp a tethered ball from a moving UAV at approximately 6 m/s.

The future work to be performed entails the improvement of the vision components, in order to provide faster and more accurate estimations of the targets. The position estimator of the FES component can be improved to decrease the position error and the detectors can be optimized to increase their frequency. Additionally, the weight has to be drastically reduced in order to enable maneuvers at even higher speeds. The exploration of different gripper designs and coupled control strategies has also been remained as future work.

\section*{Acknowledgements}
This work has been supported by the Spanish Ministry of Science, Innovation and Universities (COMCISE RTI2018-100847-B-C21, MCIU/AEI/FEDER, UE) and the Khalifa University under contract no. 2020-MBZIRC-10. The research of two authors is also supported by the Chinese Scholarship Council and the Regional Ministry of Education, Youth and Sport of the Community of Madrid and the European Social Fund, respectively.

%
%


\bibliographystyle{unsrtnat}

\end{document}